\pdfoutput=1

\documentclass[11pt]{article}

\usepackage{ACL2023}

\usepackage{times}
\usepackage{latexsym}

\usepackage{forest}
\usepackage{array}
\usepackage{textcomp}
\usepackage{amsfonts}
\usepackage{pgfplots}

\usepackage{booktabs}
\usepackage{listings}
\usepackage{pythonhighlight}

\usepackage{tcolorbox}
\usepackage{subcaption}
\usepackage{xcolor}
\usepackage{courier}
\usepackage{url}

\usepackage{multirow}
\usepackage{enumitem}
\usepackage{amsmath}

\usepackage{color,soul}
\pgfplotsset{compat=1.18} 
\usepackage[T1]{fontenc}

\usepackage[utf8]{inputenc}

\usepackage{microtype}

\usepackage{inconsolata}
\DeclareUnicodeCharacter{0307}{\.}

\title{A Taxonomy for Data Contamination in Large
Language Models}

\author{Medha Palavalli \and Amanda Bertsch \and Matthew R. Gormley\\
    School of Computer Science \\
    Carnegie Mellon University \\
    \texttt{[mpalaval, abertsch, mgormley]@cs.cmu.edu} \\
  }

\begin{document}

\maketitle
\begin{abstract}
Large language models pretrained on extensive web corpora demonstrate remarkable performance across a wide range of downstream tasks. However, a growing concern is data contamination, where evaluation datasets may be contained in the pretraining corpus, inflating model performance. 
Decontamination, the process of detecting and removing such data, is a potential solution; yet these contaminants may originate from altered versions of the test set, evading detection during decontamination. How different types of contamination impact the performance of language models on downstream tasks is not fully understood. We present a taxonomy that categorizes the various types of contamination encountered by LLMs during the pretraining phase and identify which types pose the highest risk.  We analyze the impact of contamination on two key NLP tasks---summarization and question answering---revealing how different types of contamination influence task performance during evaluation.
\end{abstract}

\section{Introduction}

Advancements in machine learning have traditionally relied on benchmark datasets to evaluate and compare model performance \cite{raji2021ai, gururaja-etal-2023-build}. With the surge of large language models (LLMs) in recent years, these benchmarks are now leveraged to showcase remarkable abilities across diverse tasks.

However, the shelf life of benchmarks is incredibly low, with \citet{roberts2023data} demonstrating that newer models with updated training cutoff dates are iteratively rendering existing benchmarks stale. 
The presence of internet-sourced data in both pretraining and evaluation datasets increases the risk of data contamination \cite{brown-2020-language, magar2022data} and challenges the notion of fair evaluation for models pretrained on massive corpora. 
Both GPT-3 and C4 training corpora were found to contain test data for several benchmarks \cite{dodge-etal-2021-documenting, raffel-etal-2020-t5, brown-2020-language}, raising serious concerns about the validity of evaluation scores for many pretrained models \cite{lee-etal-2022-deduplicating, chang2023survey}.

The research community lacks consensus on best practices for data contamination, and different works define contamination in subtly different ways. Without standardization of terminology, it is difficult to develop best practices for contamination-- or even to characterize the problem at all.
To address this gap, we suggest a formal definition of contamination and taxonomize subtypes of contamination (\S~\ref{sec:taxonomy}). We map prior work on both the detection and impact of contamination into this taxonomy, revealing several understudied forms of contamination (\S~\ref{sec:mapping_prior_work}).
We also measure the impact of different types of contamination on downstream summarization (\S~\ref{sec:summarization}) and QA (\S~\ref{sec:qa}) performance through continued pretraining experiments assessing indirect/approximate test set contamination effects. 

Our findings reveal that for GPT-2 Large models, it is often the case that having in-domain data present during training is as beneficial as having the test data present during training.
Moreover, we observe that certain contamination types exhibit task-dependent effects on evaluation performance, further complicating decontamination best practices. 
Our findings enable recommendations for identifying and mitigating problematic contamination during LLM development to ensure reliable evaluations (\S~\ref{sec:conclusion}).

\section{Taxonomy}
\label{sec:taxonomy}


\definecolor{blue}{RGB}{100, 149, 237}
%

\definecolor{skyblue}{RGB}{31, 119, 180}
\definecolor{bluishgreen}{RGB}{44, 160, 44}
\definecolor{yellow}{RGB}{188, 189, 34}
\definecolor{vermillion}{RGB}{214, 39, 40}
\definecolor{reddishpurple}{RGB}{148, 103, 189}
\definecolor{saffron}{RGB}{227, 119, 194}

\definecolor{saffron}{RGB}{153, 102, 204}

\begin{figure*}[!htb]
\footnotesize
\begin{center}
\begin{forest}
for tree={
    if level = 0 {
        rotate=90,
        parent anchor=south,
        child anchor=west
    }{},
    if level = 1 {
        text width=2.1cm,
        yshift=3.35mm,
        parent anchor=east,
        child anchor=west
    }{},
    if level = 2{
        parent anchor=east,
        child anchor=west,
        text centered,
        text width=1.75cm,
    }{},
    if n children=0{ 
        text width=7.4cm, 
    }{},
    grow'=east,
    style={rounded corners},
    edge path={
      \noexpand\path [draw, \forestoption{edge}]
      (!u.parent anchor) -- +(5pt,0) |-
      (.child anchor)\forestoption{edge label};
    },
},
[Contamination, draw=saffron, fill=saffron!15 
    [Instance Level Contamination {(\S\ref{subsubsec:instance-properties})}, draw=vermillion, fill=vermillion!10, yshift=0.15mm,
        [Augmenting, draw=blue, fill=blue!10, yshift=0.05mm, 
            [\citet{karmakar2022codex}, draw=bluishgreen!0, fill=bluishgreen!10, ]
        ],
        [Noising, draw=blue, fill=blue!10, yshift=0mm,
            [\citet{yang2023rethinking-1}, draw=bluishgreen!0, fill=bluishgreen!10, ]
        ],
        [Masking, draw=blue, fill=blue!10, yshift=0.05mm,
            [\citet{karmakar2022codex}, draw=bluishgreen!0, fill=bluishgreen!10, ]
        ],
    ],
    [Dataset Level Contamination {(\S\ref{subsubsec:dataset-properties})}, draw=vermillion, fill=vermillion!10
        [Distribution, draw=blue, fill=blue!10, yshift=-0.32mm,
            [\citet{jiang2024investigating}, draw=bluishgreen!0, fill=bluishgreen!10]
        ],
        [Selection, draw=blue, fill=blue!10, yshift=-2.1mm,
            [\citet{jiang2024investigating, zhang2023counterfactual-1, cao2024concerned, sainz-lm-contamination, li2023task},  draw=bluishgreen!0, fill=bluishgreen!10,]
        ]
    ]
]
\end{forest} 
\end{center} 
\vspace{1em}
\begin{center}
\begin{forest}
for tree={
    if level=0{
        yshift=-1.4mm,
        rotate=90,
        parent anchor=south,
        child anchor=west
    }{},
    if level=1{
        text width=3.7cm, 
        parent anchor=east,
        child anchor=west
    }{},
    if level=2{
        text width=3cm, 
        parent anchor=east,
        child anchor=west
    }{},
    if n children=0{ 
        if level=2{
            text width=8.2cm, 
            yshift=-2.14mm
        }{},
        if level=3{
            text width=6.5cm, 
        }{},
    }{},
    grow=east,
    style={rounded corners},
    edge path={
      \noexpand\path [draw, \forestoption{edge}]
      (!u.parent anchor) -- +(5pt,0) |-
      (.child anchor)\forestoption{edge label};
    },
    },
[Not Contamination, draw=saffron, fill=saffron!15
    [Transductive Learning {(\S\ref{subsubsec:trans-understanding})}, draw=vermillion, fill=vermillion!10, yshift=-0.45mm,
        [\citet{jiang2024investigating, ouchi-etal-2019-transductive, sainz-etal-2023-nlp}, draw=bluishgreen!0, fill=bluishgreen!10,]                      
    ],
    [Prior Task Understanding {(\S\ref{subsubsec:task-understanding})}, draw=vermillion, fill=vermillion!10, yshift=-0.44mm,
        [\citet{li2023task, sainz-etal-2023-nlp}, draw=bluishgreen!0, fill=bluishgreen!10]
    ],
    [Language Understanding {(\S\ref{subsubsec:lang-understanding})}, draw=vermillion, fill=vermillion!10, yshift=-0.4mm,
        [(), draw=bluishgreen!0, fill=bluishgreen!10, color=bluishgreen!10]                         
    ]
]
\end{forest}
\end{center}
\caption{Taxonomy of Contamination, with some representative works in the literature that address each category.}
\vspace{-1.1em}
\label{fig:taxo-human-feedback}
\vspace{1.1em}
\end{figure*}
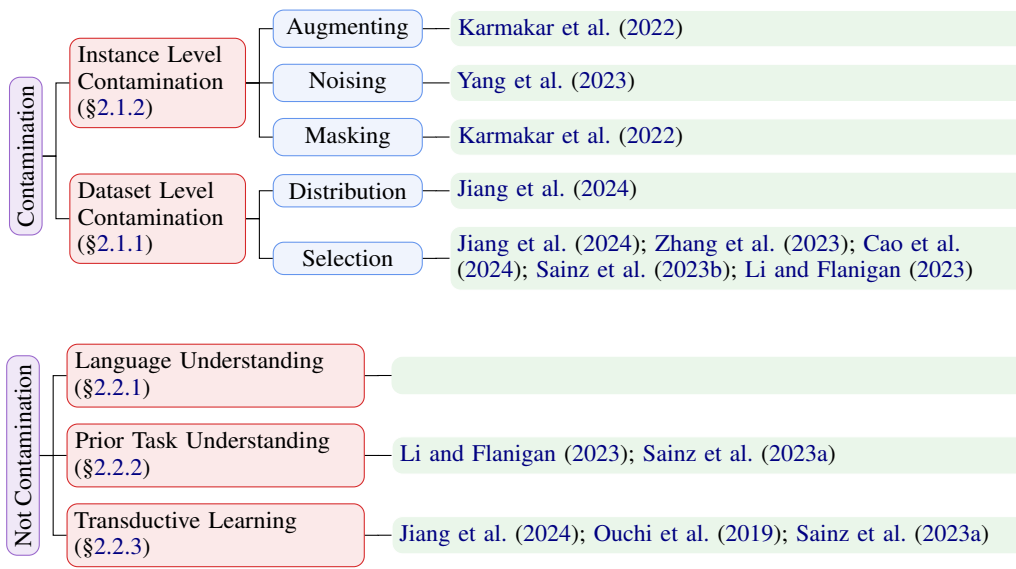

Consider a model $M : \mathcal{X} \rightarrow \mathcal{Y}$ which, given an input of some type $x \in \mathcal{X}$, outputs text $\widehat{y} \in \mathcal{Y}$. While $x$ can be of any format, we will restrict ourselves to cases where $\widehat{y}$ is in the space of the \textit{natural language} ($\mathcal{Y} \subseteq \Sigma^*$ for some alphabet in $\Sigma$). Let $D$
be the \textit{test} set, consisting of $|D|$ examples $\langle x_i, \widehat{y}_i \rangle$. 

\subsection{Contamination}
\label{subsec:contamination}
We define \textbf{contamination} as any leakage of information that provides a signal for the correct label for at least one example in the test set $D$.  When contamination occurs, some subset of the pretraining data can be characterized as the result of a function $f(D)$, which may be a composition of multiple contamination functions 
$f = f^{(1)} \circ f^{(2)} \circ \cdots \circ f^{(n)}$.
 We characterize types of contamination by their \emph{dataset-level} (\S~\ref{subsubsec:dataset-properties}) and \emph{example-level} (\S~\ref{subsubsec:instance-properties}) properties. Figure \ref{fig:taxo-human-feedback} provides an overview of our taxonomy.

\subsubsection{Dataset-level Properties}
\label{subsubsec:dataset-properties}
For dataset-level contamination, consider a function $g$ that leaves the individual examples $\langle x_i, \widehat{y}_i \rangle$ intact. In the simplest case, $g$ is the identity function; this is the leakage of a full test set, e.g. from scraping a file containing the test set instances and labels. The following are types of functions $g(D)$ can take on.
\begin{itemize}
    \item \textbf{Selection:} A function that selects some group of examples $D' \subset D$, such that only a subset of the test set is leaked. This is likely when the test data is drawn from several sources, only some of which appear in the pretraining data; when some of the test data is more recent than other data and the pretraining data contains an older snapshot of the contamination source; or when the data is contained in several documents and the cleaning of the pretraining data only removes some of these documents. \emph{Verbatim contamination} refers to when $g$ is the identity function.
    \item \textbf{Distribution:} A function which combines the contaminated data $D$ with some additional, non-contaminating documents, such that the examples from $D$ are not all sequential in the pretraining data. This can occur during data shuffling, or if the contamination comes from multiple documents. Practically, this means that the contaminated region of the pretraining data $g(D)$ spans more tokens.
\end{itemize}

\subsubsection{Instance-level Properties} 
\label{subsubsec:instance-properties}
In instance-level contamination, the function $f$ applies some function $h$ to each individual leaked example $f(D) = \{ h(\langle x_i, \widehat{y}_i \rangle ) \}_{i=1}^{|D|}$.\footnote{Note that this is a strict subset of all functions applied to the leaked dataset, $f(D)$; however, we distinguish this set of functions that operate on individual examples.} A few representative examples in this class are enumerated below:
\begin{itemize}
    \item \textbf{Masking:} A function that removes some or all of the input (can be done in combination with the output), e.g. $h( \langle x_i, \widehat{y}_i \rangle) = \widehat{y}_i$ or removing all incorrect answer choices in a multiple choice question. \emph{This primarily qualifies as contamination for generation tasks;} for a classification task, leaking the label-space in advance may not be a concern if the labels don't have inherent contextual value without the input, such as binary labels like 0s and 1s or positives and negatives. However, if the labels carry meaningful information on their own, their premature disclosure would indeed constitute contamination. Note that masking \emph{all of the output}, leaving only the inputs from the test set, is generally considered to be a type of transductive learning, \textit{not} contamination; see \S~\ref{subsubsec:trans-understanding} for more discussion.
    \item \textbf{Noising:} A function that modifies the surface form of the example, e.g. by paraphrasing the inputs or outputs, by presenting the output before the input, or by using silver rather than gold labels for each example. Note that this can also take the form of \textit{alternate correct answers} being present in the pretraining data: for instance, in book summarization, a different summary of the book being present in the pretraining data is still contamination. 
    \item \textbf{Augmenting:} A function that adds additional context, which may or may not be relevant to the example. For instance, for a task where the model must answer an open-ended question at test time, an \textit{augmented} contaminated example in pretraining would be a multiple-choice test with the same questions. While this provides the correct answer, it also introduces new (distractor) information that is not present at test time. Another example would be including additional context paragraphs for QA in addition to the necessary context and answer. Note the difference between example-level \emph{augmenting} and dataset-level \emph{distribution}.
\end{itemize}

\subsection{Phenomena that aren't Contamination}
\label{subsec:not-contamination}
For clarity, we describe several phenomena that lead to improved performance on test sets downstream but are \textit{not} considered contamination under our taxonomy.\\

\subsubsection{Language Understanding}
\label{subsubsec:lang-understanding}
Pretraining enables models to produce (generally) fluent text and encodes some representation of meaning for words commonly used in task definitions; for instance, the model has some representation of meaning for the labels ``positive'' and ``negative'' in sentiment analysis. While this representation is likely helpful for performing downstream tasks \cite{min-etal-2022-rethinking}, this is not inherently contamination.\\

\subsubsection{Prior Task Understanding}
\label{subsubsec:task-understanding}
We define prior task understanding as an ability to perform a task learned from non-contamination sources, and such prior knowledge has been demonstrated to boost model performance when evaluated on unseen instances of said task \cite{li2023task}. For instance, fine-tuning a model on a training dataset for the task is clearly not contamination of the test set, although it generally improves performance on that test set; likewise, pretraining on other related datasets is not contamination for a given test set. For closed-book QA and tasks requiring world knowledge, prior task understanding from training data is essential. Closed-book QA demands answering without external resources, relying solely on the model's training on similar question-answer pairs or related datasets.

In general, scrutinizing the training data's sources and nature is crucial to maintain model integrity and generalizability.
Prior task understanding may violate the assumption of ``zero-shot'' performance: that the model has not seen training data for that task.

\subsubsection{Transductive Learning}
\label{subsubsec:trans-understanding}
Transductive learning \cite{vapnik-transductive} incorporates an unlabeled test set into training. During training, the raw text inputs of the test set can be used, but the labels are not seen. The model, once trained, is then evaluated on the same test set during the test phase. 
Transductive LM fine-tuning has shown to consistently improve neural models in both in-domain and out-of-domain settings \cite{ouchi-etal-2019-transductive}, although concerns have been raised about blurring the line between training and evaluation \cite{jiang2024investigating}.

We generally do not consider pretraining on the \textit{inputs} of the test set to be contamination,\footnote{A key exception is tasks where the input/output distinction does not apply, such as perplexity evaluation on a dataset $D = \{ x_1, \ldots, x_{|D|}\}$ of sentences $x_i$.} although we note that this will likely improve performance, in the same manner than pretraining on training set text improves downstream performance by providing some domain adaptation to the testing domain \cite{gururangan-etal-2020-dont, krishna2023downstream}. 
Some prior work refers to the presence of inputs-only in the pretraining data as contamination for classification tasks \cite{jiang2024investigating, ouchi-etal-2019-transductive}; however, under our taxonomy, we consider this a type of transductive learning. 

\subsection{Mapping prior work exploring contamination into this taxonomy}
\label{sec:mapping_prior_work}

The effects of \emph{selection} have been explored by experiments that compare LLM performance over time \cite{li2023task, cao2024concerned}, prompting the model to generate samples from specific dataset splits \cite{sainz-lm-contamination}, and training LLMs that \emph{select} some subset of an evaluation dataset \cite{zhang2023counterfactual-1, jiang2024investigating}.

\citet{jiang2024investigating} also explores the effects of the frequency in which contaminated data appears \emph{distributed} throughout the pretraining data. 

Through zero-shot experimentation on the Codex model \cite{chen2021evaluating}, \citet{karmakar2022codex} investigates the effects of prompts \emph{masking} out input specifications and prompts with \emph{augmented} objectives. Additionally, \citet{yang2023rethinking-1} showcases memorization of evaluation samples by prompting LLMs with \emph{noisy} samples.

A prior position paper \cite{sainz-etal-2023-nlp} defined three categories of data contamination: their \textit{guideline contamination} falls under our definition of prior task understanding; their \textit{raw text contamination} is tranductive learning; and their \textit{annotation contamination} equates to our definition of data contamination in \S~\ref{subsec:contamination}. Our work further categorizes and explores types of annotation contamination.

\subsection{Detecting Data Contamination}

\textbf{Methods with access to pretraining data}
Early research on LLM data contamination primarily employed methods akin to high-order n-gram overlap detection between pretraining and evaluation data \cite{radford-gpt-eval, brown-2020-language, wei-zero-shot, touvron-2023-llama}. Tools for qualitative analysis on large-scale corpora (such as Data Portraits \cite{marone2023data} and the ROOTS Search Tool \cite{piktus2023roots}) have further increased the practicality of this type of contamination detection. 
However, these approaches have several limitations: they remain fairly computationally expensive, assume access to pretraining data, and generally can only detect contamination when a cluster of several test set examples co-occur (as most methods leverage data sketching \cite{broder-datasketch} tools that are only effective for sequences above a certain length). 

\citet{yang2023rethinking-1} proposes an LLM-based decontamination method, which leverages embedding similarity search followed by evaluation with a strong language model (e.g. GPT-4), to identify and mitigate contamination. This is computationally costly but can identify \emph{noisy} contamination

\textbf{Methods without access to pretraining data}
Some approaches are capable of detecting contamination without direct access to pretraining data, but assume that the test data has not been modified or distributed across the pretraining corpus. These methods leverage \textit{metadata} from the dataset to detect contamination, e.g. by leveraging dataset ordering \cite{sainz-lm-contamination} or the assignment of examples to specific data splits \cite{ golchin2023time}. \citet{golchin2024datacontaminationquiztool} introduce the Data Contamination Quiz, a streamlined method that efficiently detects and
estimates verbatim contamination in LLMs by crafting multiple choice questions that prompt a model to correctly dataset-specific content among similar but noisy alternatives.

\citet{chang2023speak} detect contamination of books (which serve as inputs for many long-context evaluation datasets) using domain specific features-- a name cloze test and a publication-year evaluation. This is powerful for detecting the presence of the exact text of the book, but its efficacy on detecting related artifacts (e.g. summaries of the book, which may serve as test set outputs) is unknown.

\citet{shi2023detecting} introduces a new detection method MIN-K\% PROB, which is capable of detecting whether a piece of text was in the pretraining corpora by leveraging the variability of the tokens' probabilities according to the model. This has the potential to detect distributed or masked contamination, but is not robust to noising operations, which change the token sequence.

Most contemporary data-contamination detection techniques are designed to identify contamination of full, non-distributed test datasets, resulting in a significant gap in detecting noisy or partial contamination. The methods most well-adapted to detect noisy contamination, while powerful, require access to pretraining data and expensive operations; more work is necessary to lower the barrier to detection.

\section{Methodology}
In all our experiments, we employ GPT-2 Large \cite{radford2019language}.\footnote{Our implementation uses nanoGPT \cite{nanoGPT} initialized with OpenAI’s \texttt{gpt2-large} weights.} This will be referred to as the \emph{initial model}. Since the pretraining corpus for GPT-2 is not publicly accessible, there is a chance that these \emph{learned} weights of GPT-2 might be contaminated. Consequently, the outcomes of our experiments serve as a conservative estimate or lower bound on the effects of data contamination.

For each of our datasets, we create \texttt{train}/\texttt{in-domain}/\texttt{test} splits of equal size, aiming to establish a fair and comparable evaluation environment. To disentangle the effects of exposure to \texttt{test} data during pretraining from those of prior task understanding, we constructed an \texttt{in-domain} data split, allowing us to train models on task-relevant but uncontaminated data for comparison against the various contaminated settings. To partially mitigate the potential recency bias from continued pretraining, we incorporate an additional 10,000 samples of Open AI's WebText \cite{radford2019language} into the continued pretraining data.

During continued pretraining, we use a blocksize of 1024 tokens with a batchsize of 1. For finetuning, the training data is seen sample by sample. To obtain deterministic results during our experiments, we set the temperature to zero and capped the maximum completion length at 200 tokens.

\subsection{Training Settings}

We consider several settings for incorporating data:
\begin{itemize}[itemsep=0pt,parsep=0pt]
    \item \textsc{zero-shot} (not contamination): prompt the initial model with the test sample and a simple instruction for the task. 
    \item \textsc{baseline} (not contamination): finetune initial model with \texttt{train} split
    \item \textsc{cheating} (contamination at fine-tuning time, rather than pre-training): finetune initial model with \texttt{test} split
    \item \textit{Contamination Setting(s)} (standard contamination during pretraining): continued pretraining with $f$(\texttt{test} split) and finetune with \texttt{train} split; the details of each contamination setting are specific to the task (\S~\ref{sec:summarization} and \S~\ref{sec:qa})
    \item \textit{In-Domain Setting(s)} (not contamination): continued pretraining with $f$(\texttt{in-domain} split) and finetune with \texttt{train} split---for each contamination setting in \S~\ref{sec:summarization} and \S~\ref{sec:qa}, there is an associated in-domain model.
\end{itemize}
For each setting+dataset, we average results over models trained on 3 random shuffles of the data. Standard deviations are computed over these 3 runs and error bars indicate $\pm$ one standard deviation.

\section{Case Study: Summarization}
\label{sec:summarization}
\definecolor{vermillion}{RGB}{214, 39, 40}
\definecolor{bluishgreen}{RGB}{44, 160, 44}
\definecolor{blue}{RGB}{100, 149, 237}

\begin{figure*}
\centering
\begin{tikzpicture}
\begin{axis}[
    ybar,
    width=\textwidth, 
    height=6cm, 
    bar width=15pt,
    ylabel={Rouge-L},
    xlabel style={font=\small, yshift=-0.75cm},
    area style,
    xtick=data,
    ymax=29.5,
    ymin=24,
    xticklabels={\textsc{baseline},  \textsc{distribution}, \textsc{masked}, \textsc{noised}, \textsc{reformatted}, \textsc{verbatim}, \textsc{cheating}},
    xticklabel style={align=center, font=\tiny, text width=3cm}, 
    legend style={at={(0.5,0.2)}, anchor=north,legend columns=-1, font=\footnotesize},
    nodes near coords, 
    nodes near coords align={vertical},
    nodes near coords style={font=\tiny, yshift=8pt},
    point meta=explicit symbolic,
    ]

    \addplot[fill=black!0, draw=black!70, point meta=explicit symbolic, bar width=0pt] coordinates {
        (2, 0)
        (3, 0)
        (4, 0)
        (5, 0)
        (6, 0)
        (7, 0)
        (8, 0)
    };
    
    \addplot[fill=bluishgreen!20, draw=bluishgreen!70, xshift=7.5pt, error bars/.cd,y dir=both,y explicit] coordinates {
        (3, 26.32)  +- (0, 0.33)
        (4, 25.74)  +- (0, 0.35)
        (5, 26.38)  +- (0, 0.29)
        (6, 26.24)  +- (0, 0.28)
        (7, 26.50)  +- (0, 0.30)
    };

    \addplot[fill=black!0, draw=black!70, error bars/.cd,y dir=both,y explicit] coordinates {
        (2, 25.83) +- (0, 0.32) [6]
        (8, 28.41) +- (0, 0.33) [1]
        
    };

    \addplot[fill=vermillion!20, draw=vermillion!70, xshift=-7.5pt, error bars/.cd,y dir=both,y explicit] coordinates {
        (3, 26.48) +- (0, 0.31) [3]
        (4, 25.77) +- (0, 0.30) [7]
        (5, 26.39) +- (0, 0.37) [4]
        (6, 26.37) +- (0, 0.31) [5]
        (7, 26.98) +- (0, 0.40 )[2]
    };
    
   \addplot[fill=black!0, draw=black!70, point meta=explicit symbolic, bar width=0pt] coordinates {
        (2, 0)
        (3, 0)
        (4, 0)
        (5, 0)
        (6, 0)
        (7, 0)
        (8, 0)
    };

     \legend{,In-Domain, ,Contamination}
    
\end{axis}
\end{tikzpicture}
\caption{Bar Chart of all SAMSum models compared for Rouge-L.}
\label{fig:summ_sample}
\end{figure*}
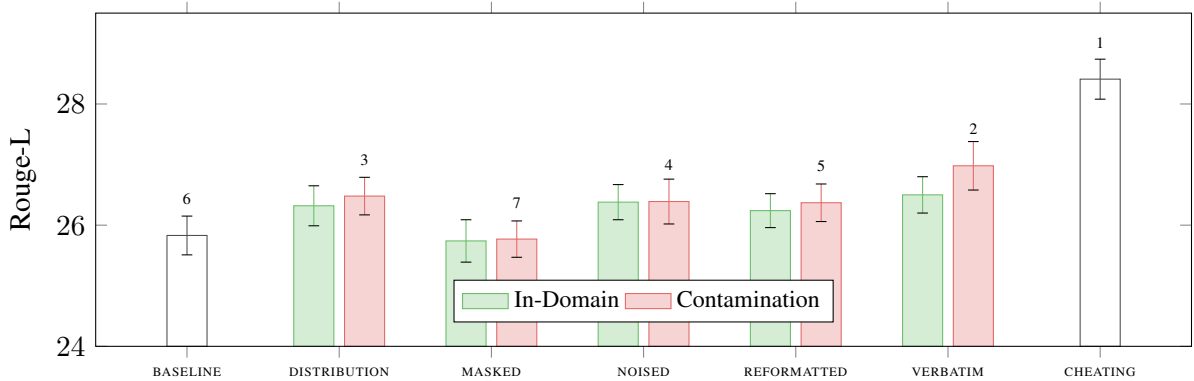

For this case study, we use the following summarization datasets: XSum \cite{narayan-etal-2018-dont}, SAMSum \cite{gliwa-etal-2019-samsum}, and CNN/Daily Mail \cite{nallapati-etal-2016-abstractive}. 
We explore 5 contamination settings:
\begin{enumerate}[itemsep=0pt,parsep=0pt]
    \item \textsc{verbatim} (dataset level, selection): $f = $ identity function on \texttt{test} split
    \item \textsc{distribution} (dataset level, distribution): $f = $ shuffle \texttt{test} data with WebText
    \item \textsc{masked} (instance level, masking): $h = $ mask out input documents in \texttt{test} split
    \item \textsc{noised} (instance level, noising): $h = $ swap in GPT-3.5\footnote{\texttt{gpt-3.5-turbo-0125} with temperature=0.5} generated summaries on \texttt{test} split
    \item \textsc{reformatted} (instance level, noising): $h = $ swap format from document-summary to summary-document for \texttt{test} data
\end{enumerate}
Table \ref{tab:summ_example} provides examples of each setting.

\subsection{Results}
In this section, we consider the overall performance of each contamination method across summarization datasets. Figure \ref{fig:summ_sample} shows an example of the results from one task and one metric (SAMSum, ROUGE-L). See Appendix \ref{app:full_results_summ} for full results on all tasks and metrics, specifically Figures \ref{fig:cdm_barchart}, \ref{fig:samsum_barchart}, \ref{fig:xsum_barchart} or Table \ref{tab:full_summ}.

Consistently, the \textsc{cheating} setting outperforms all others; this is expected, given that deliberately finetuning on the \texttt{test} data is an extreme form of contamination. 

Overall, continued pretraining with the approximate contamination methods improves performance above the \textsc{baseline} setting, often substantially. This suggests that exposure to these forms of contamination during pretraining can impact the reliability of evaluations on this data downstream. 

While \textsc{verbatim} setting performs slightly better than the other contamination settings, this improvement isn't significant for most settings. Note that most contaminated settings outperform the baseline, and exist within a standard deviation of each other. This suggests that the performance boost may simply be attributed to the increase in \texttt{in-domain} data seen during the training stage rather than encountering the \texttt{test} split during continued pretraining.

Note that for the most part, the \textsc{verbatim} and \textsc{indomain-verbatim} settings perform on par with each other. This trend seems to hold true for the other \emph{contamination} and \emph{in-domain} model pairs. The comparable performance further suggests that exposure to contaminated data may not be the primary factor boosting model performance in the \emph{contamination} settings studied.

\begin{table} [ht]
  \centering
  \begin{tabular}{ccccc}
    \toprule
    Dataset & R-1 & R-2 & R-L& R-Lsum\\
    \midrule
    CNN & 38.70 & 14.14 & 24.90 & 32.11 \\
    SAMSum & 37.92 & 13.78 & 28.73 & 28.75 \\
    XSum & 24.21 & 4.89 & 16.60 & 16.60 \\
    \bottomrule
  \end{tabular}
  \caption{Rouge scores (R-) for summaries generated by GPT-3.5. These summaries are used as silver labels for our \textsc{noised} contamination setting.}
  \label{tab:summnoise}
\end{table}

While the majority of these settings have metrics that fall within one standard deviation of each other, there are exceptions. For instance, in the case of the XSum dataset, the \textsc{noised} setting fails to surpass the \textsc{baseline}. This discrepancy can be attributed to the idiosyncrasies of the XSum dataset, where ground truth summaries may deviate significantly from typical summaries, thus posing a challenge for the model in generating accurate outputs. Table \ref{tab:summnoise} shows that the summaries generated by GPT-3.5 \cite{brown-2020-language} for the XSum dataset have lower rouge scores than the other two datasets.

Additionally, underperformance of the \textsc{masked} contamination setting compared to the \textsc{baseline} across all datasets is noteworthy, suggesting that exposure only to summaries during pretraining may fail to achieve the benefits of seeing \texttt{in-domain} data.

\section{Case Study: Question Answering}
\label{sec:qa}

For this case study, we consider open-ended QA with SQuAD \cite{rajpurkar-etal-2016-squad} and multiple-choice QA with the Children's Book Test (CBT) \cite{hill2016goldilocks}. 
We explore 6 contamination settings:
\begin{enumerate}[itemsep=0pt,parsep=0pt]
    \item \textsc{verbatim} (dataset level, selection): $f = $ identity function on \texttt{test} split
    \item \textsc{distribution} (dataset level, distribution): $f = $ shuffle \texttt{test data} with WebText
    \item \textsc{masked} (instance level, masking): $h = $ mask out context passage in \texttt{test} split
    \item \textsc{noised} (instance level, noising): $h = $ encounter GPT-3.5 generated answers to \texttt{test split} questions
    \item \textsc{reformatted} (instance level, augmenting/masking)\footnote{For SQuAD, this is a form of augmented contamination, as additional (distractor) information is introduced. For CBT, this is a form of masked contamination, as information is removed.}: $h_{\text{SQuAD}} = $ introduce 3 distractor multiple choice answer options; $h_{\text{CBT}} = $ mask out incorrect answer options
    \item \textsc{augmented} (instance level, augmenting):
    $h = $ prompt GPT-3.5 to add additional content to the context passages in the \texttt{test} split
\end{enumerate}
Table \ref{tab:qa_example} provides examples of each setting.

\definecolor{vermillion}{RGB}{214, 39, 40}
\definecolor{bluishgreen}{RGB}{44, 160, 44}
\definecolor{blue}{RGB}{100, 149, 237}

\begin{figure*}
\centering
\begin{tikzpicture}
\begin{axis}[
    ybar,
    width=\textwidth, 
    height=6cm, 
    bar width=15pt,
    ymin=37,
    ymax=58.5,
    ylabel={Exact Match},
    xlabel style={font=\small, yshift=-0.75cm},
    area style,
    xtick=data,
    xticklabels={\textsc{baseline},  \textsc{distribution}, \textsc{masked}, \textsc{noised}, \textsc{reformatted}, \textsc{augmented}, \textsc{verbatim}, \textsc{cheating}},
    xticklabel style={align=center, font=\tiny, text width=3cm}, 
    legend style={at={(0.62,0.2)}, anchor=north,legend columns=-1, font=\footnotesize},
    nodes near coords, 
    nodes near coords align={vertical},
    nodes near coords style={font=\tiny, yshift=4pt},
    point meta=explicit symbolic,
    ]

    \addplot[fill=black!0, draw=black!70, point meta=explicit symbolic, bar width=0pt] coordinates {
        (2, 0)
        (3, 0)
        (4, 0)
        (5, 0)
        (6, 0)
        (7, 0)
        (8, 0)
        (9, 0)
    };
    
    \addplot[fill=bluishgreen!20, draw=bluishgreen!70, xshift=7.5pt, error bars/.cd,y dir=both,y explicit] coordinates {
        (3, 51.895) +- (0, 0.91)
        (4, 44.615) +- (0, 0.93)
        (5, 50.625) +- (0, 0.85)
        (6, 51.3) +- (0, 0.95)
        (7, 52.94) +- (0, 0.94)
        (8, 52.442) +- (0, 0.89)
    };

     \addplot[fill=black!0, draw=black!70, error bars/.cd,y dir=both,y explicit] coordinates {
        (2, 41.768) +- (0, 1.01) [7]
        (9, 55.731) +- (0, 0.94) [1]
    };

    \addplot[fill=vermillion!20, draw=vermillion!70, xshift=-7.5pt, error bars/.cd,y dir=both,y explicit] coordinates {
        (3, 52.768) +- (0, 0.89) [4]
        (4, 38.779) +- (0, 0.96) [8]
        (5, 52.726) +- (0, 0.89) [5]
        (6, 48.081) +- (0, 0.91) [6]
        (7, 53.589) +- (0, 0.98) [2]
        (8, 53.389) +- (0, 0.94) [3]
    };

    \addplot[fill=black!0, draw=black!70, point meta=explicit symbolic, bar width=0pt] coordinates {
        (2, 0)
        (3, 0)
        (4, 0)
        (5, 0)
        (6, 0)
        (7, 0)
        (8, 0)
        (9,0)
    };
    
    \legend{,In-Domain, ,Contamination}
    
\end{axis}
\end{tikzpicture}

\caption{Bar Chart of all SQuAD models compared for Exact Match.}
\label{fig:qa_sample}
\end{figure*}
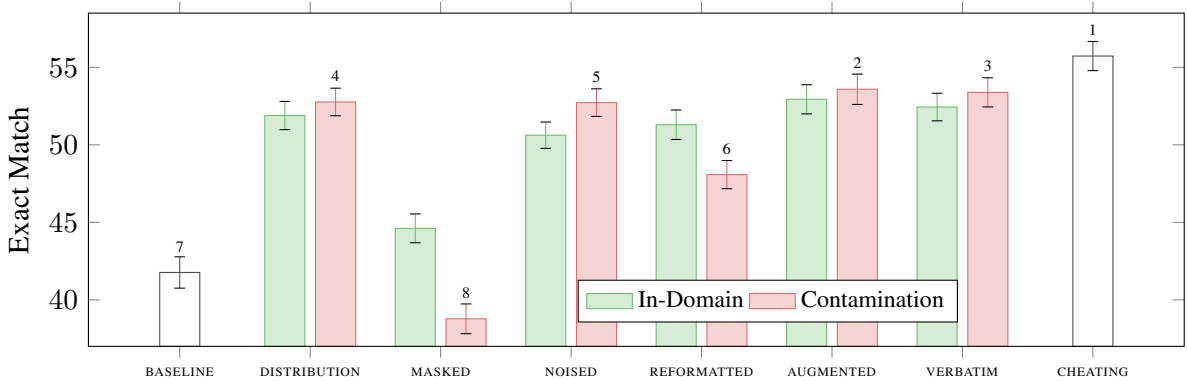
\subsection{Results}
In this section, we consider the overall performance of each contamination method across
Question Answering datasets. Figure \ref{fig:qa_sample} shows an example of the results from one task and one metric (SQuAD, Exact Match). See Appendix \ref{app:full_results_qa} for full evaluation results on all tasks and metrics, specifically Figures \ref{fig:squad_barchart}, \ref{fig:cbt_barchart} or Table \ref{tab:full_qa}.

Once again, the \textsc{cheating} setting outperforms all others by a noticeable margin. With the exception of the \textsc{masked} setting for the SQuAD dataset, all contaminated settings exhibit better performance compared to the \textsc{baseline} setting by a considerable margin. This indicates that the increased data diversity experienced by both the \emph{in-domain} and \emph{contaminated} models during training improved their performance during evaluation. 

\begin{table}[ht]
  \centering
  \begin{tabular}{ccc}
    \toprule
    Dataset & Exact Match & F1 Score\\
    \midrule
    SQuAD & 74.56 & 88.15\\
    CBT & 77.21 & 79.78\\
    \bottomrule
  \end{tabular}
  \caption{Exact match and F1 scores for answers generated by GPT-3.5.}
  \label{tab:qanoise}
\end{table}

Note that the \textsc{noised} setting performs almost as well as the \textsc{verbatim} contamination setting. We attribute this to the fairly high quality of silver labels generated by GPT-3.5 (see Table \ref{tab:qanoise}).

Exposure to \texttt{in-domain} data during pretraining appears to improve model performance. However, our results show that contaminated settings such as \textsc{noised}, \textsc{verbatim}, and \textsc{distribution} tend to outperform the corresponding \emph{in-domain} settings during evaluation. This suggests that seeing data from the \texttt{test} set positively impacts model performance for question answering tasks. Note that for these three model setups, the format of context, question, and answer is almost consistent with the format and content seen during evaluation time.

Reformatting (augmenting) free-form questions from SQuAD into multiple-choice answers during pretraining appears to have a negative effect on model performance, though it still outperforms the \textsc{baseline} setting. Conversely, converting multiple-choice questions from CBT into free-form questions (masking) during pretraining yields positive results, with the \textsc{reformatted} setting outperforming most other contaminated settings.

Furthermore, we observe variations in the performance of \textsc{augmented} setting across the two datasets. While this setting perform well for SQuAD, its performance is not as impressive for CBT. This discrepancy may be attributed to the nature of data augmentation, where the additional information provided for SQuAD is more relevant and beneficial to the wikipedia paragraphs compared to the irrelevant introductions, such as `once upon a time' style introductions generated by GPT-3.5 for these book excerpts, added to CBT stories. It is important to note that since this information doesn't significantly contribute to the task, this form of augmentation falls in a blurry space between distribution and augmentation branches of the taxonomy. It could also be viewed as unrelated information being added between samples during pretraining, complicating its categorization.

\section{Analysis}
\label{sec:analysis}
Unsurprisingly, the \textsc{cheating} and \textsc{verbatim} contamination settings consistently outperform the \textsc{baseline} across both tasks. The \emph{in-domain} settings' consistent outperformance of the \textsc{baseline} underscores the advantages of exposure to related samples during pretraining \cite{krishna2023downstream}.

Far more concerning is that several approximate contamination settings outperform both the \textsc{baseline} and their respective \emph{in-domain} settings, suggesting that the model in these settings benefits not only from seeing in-domain text but from unfairly leveraging prior knowledge of the test examples. In particular, the \textsc{noised} setting, which is generally not detectable with existing decontamination methods, produces scores inflated over \textsc{baseline} in all datasets, and scores more than one standard deviation above its corresponding \emph{in-domain} setting in several datasets.

The \textsc{masked} setting generally performs around or worse than the \textsc{baseline}, possibly due to the more extreme formatting mismatch between this data and the test data. We expect that the \textsc{masked} setting may be encountered in the wild if a file of outputs for the dataset is in the pretraining data; the limited impact of this contamination on downstream performance is thus good news, though more investigation would be necessary to conclusively say \textsc{masked} contamination is not a concern. 

For many of the \emph{contaminated} settings and their corresponding \emph{in-domain} settings, the effect of approximate contamination is not greater than affect of \texttt{in-domain} data seen during pretaining. However, research has shown that memorization in LLMs significantly grows as the size of the model increases \cite{carlini2023quantifying}. The number of times a sample has been duplicated in the pretraining corpora has also been shown to increase a model's memorization capabilities \cite{carlini2023quantifying, golchin2023time}.

Some behavior is task- or dataset-specific, emphasizing that there is no one-size-fits-all approach to data curation: the importance of removing each type of contamination from the pretraining corpus is at least partially linked to the specific task's formatting. However, some types of approximate contamination do lead to inflated scores, emphasizing that considering a more broad definition of contamination when de-contaminating pretraining corpora is a worthwhile endeavor.

\section{Conclusion}
\label{sec:conclusion}

Our analysis highlights the importance of data format, with models performing better when pretraining data matches the evaluation format. We also observe task-specific effects, with certain contamination methods benefiting particular tasks more than others. Additionally, we find that some late-stage pretraining contamination can actually be \textit{unhelpful} to downstream performance, if it occurs in a substantially different format from the downstream task.
Our findings underscore gaps in current decontamination practices, which primarily focus on full-dataset-level contamination and are often unable to detect approximate or noisy contamination.

We demonstrate that different types of contamination can have variable effects on model performance, highlighting the need for careful consideration during training and evaluation. With the creation of our taxonomy, we hope to promote standardization regarding the definition and categories of contamination within the research community, facilitating clear communication and collaboration, while also enabling precise detection and mitigation of contamination in pretraining data. We recommend researchers decontaminating pretraining corpora for LLMs prioritize developing techniques that address noisy evaluation data, while also ensuring rigorous scrutiny to prevent any shuffled or interleaved evaluation data from inadvertently persisting in the pretraining data. It is not enough to merely remove instances of the full test dataset in the pretraining corpus; fragments or noised versions of the test set can also inflate performance. We hope our work inspires future work on detecting and mitigating specific types of contamination. 

\section{Limitations}
Due to resource constraints, we only investigate the impact of encountering contaminated data towards the end of pretraining (i.e. with continued pretraining), rather than randomly throughout pretraining. This may introduce recency bias, influencing our findings. Additionally, our focus on a single language model limits the generalizability of our results. GPT-2 pretraining data is not publicly accessible so our results may only offer an approximation of contamination effects. Different model architectures, training procedures, and datasets may yield varying impacts of contamination. Conducting experiments on larger LLMs could potentially reveal more pronounced effects of contamination, as larger models have been shown to exhibit greater tendencies of memorization \cite{carlini2023quantifying}. Further research involving multiple models and comprehensive evaluations is needed to establish more robust conclusions across diverse settings.

\section*{Acknowledgements}
The authors would like to thank Lori Levin for her early guidance and the anonymous reviewers for their thoughtful comments. This work was supported in part by grants from 3M, the Pittsburgh Supercomputing Center, and the National Science Foundation Graduate Research Fellowship under Grant No. DGE2140739.

\bibliography{anthology,custom}
\bibliographystyle{acl_natbib}

\appendix

\clearpage
\onecolumn

\section{Full results for Summarization Case Study}
We present the full results of the summarization case study. For each setting and dataset, we have included a table of the Rouge metrics along with their standard deviations. The data is also presented through a series of bar charts for easier interpretability of the results for the reader. Standard deviations are measured over the results of the 3 models trained on random shuffles of the data.
\label{app:full_results_summ}

\begin{table*}[ht]
    \centering
    \footnotesize
    \resizebox{\textwidth}{!}{
    \begin{tabular}{>{\arraybackslash}p{1.2cm}|>{\centering\arraybackslash}p{3.0cm}|>{\centering\arraybackslash}p{1.8cm}|>{\centering\arraybackslash}p{1.8cm}|c|c|c|c|cr}
        \toprule
        \multirow{3}{*}{Dataset} & \multirow{3}{*}{Model} & Contaminated & Contaminated & \multirow{3}{*}{ROUGE-1} & \multirow{3}{*}{ROUGE-2} & \multirow{3}{*}{ROUGE-L}  & \multirow{3}{*}{ROUGE-LSUM} \\
        & & Pretraining & Fine-tuning & & & & \\
        & & Data & Data & & & & \\
        \midrule
        \multirow{13}{*}{CNN} & \textsc{zero-shot} & - & - & 21.98 $\pm$  0.26 & 5.076 $\pm$ 0.01& 13.63 $\pm$ 0.10 & 18.51 $\pm$ 0.10\\
        & \textsc{baseline} & - & \text{\texttimes} & 27.22 $\pm$ 0.53& 7.436 $\pm$ 0.13 & 18.15 $\pm$ 0.43 & 24.90 $\pm$ 0.36 \\
        & \textsc{cheating} & - & \text{\checkmark} & \textbf{33.60} $\pm$ 0.58 & \textbf{10.198} $\pm$ 0.16 & \textbf{20.52} $\pm$ 0.33 & \textbf{29.61} $\pm$ 0.32\\
        \cline{2-8}
        \rule{0pt}{2.5ex}
        & \textsc{verbatim} & \text{\checkmark} & \text{\texttimes} & 29.84 $\pm$ 0.48 & 9.488 $\pm$ 0.14 & 19.50 $\pm$ 0.38 & 26.98 $\pm$ 0.40\\
        & \textsc{distribution} & \text{\checkmark} & \text{\texttimes} & 29.73 $\pm$ 0.33 & \textit{9.557} $\pm$ 0.13 & \textit{19.50} $\pm$ 0.22 & 27.12 $\pm$ 0.26 \\
        & \textsc{masked} & \text{\checkmark} & \text{\texttimes} & 28.34 $\pm$ 0.22& 8.326 $\pm$ 0.13 & 18.01 $\pm$ 0.29 & 25.96 $\pm$ 0.20\\
        & \textsc{noised} & \text{\checkmark} & \text{\texttimes} & \textit{31.31} $\pm$ 0.52 & 8.821 $\pm$ 0.15& 19.19 $\pm$ 0.32 & \textit{28.85} $\pm$ 0.30 \\
        & \textsc{reformatted} & \text{\checkmark} & \text{\texttimes} & 29.21 $\pm$ 0.28 & 8.887 $\pm$ 0.13 & 18.88 $\pm$ 0.29 & 26.27 $\pm$ 0.30 \\
        
        \cline{2-8}
        \rule{0pt}{2.5ex}
        & \textsc{indomain-verbatim} & \text{\texttimes} & \text{\texttimes} & 29.81 $\pm$ 0.48 & 9.277 $\pm$ 0.13 & 18.93 $\pm$ 0.25 & 26.88 $\pm$ 0.31 \\
        & \textsc{indomain-dist.} & \text{\texttimes} & \text{\texttimes} & 28.86 $\pm$ 0.30 & 8.910 $\pm$ 0.13 & 18.41 $\pm$ 0.27 & 26.10 $\pm$ 0.30\\
        & \textsc{indomain-mask} & \text{\texttimes} & \text{\texttimes} & 28.87 $\pm$ 0.39 & 8.493 $\pm$ 0.15 & 18.24 $\pm$ 0.30 & 26.40 $\pm$ 0.29\\
        & \textsc{indomain-noise} & \text{\texttimes} & \text{\texttimes} & 31.16 $\pm$ 0.42& 8.596 $\pm$ 0.10 & 18.85 $\pm$ 0.26 & 26.53 $\pm$ 0.35\\
        & \textsc{indomain-reform.} & \text{\texttimes} & \text{\texttimes} & 28.80 $\pm$ 0.31 & 8.681 $\pm$ 0.12 & 18.75 $\pm$ 0.24 & 26.07 $\pm$ 0.32\\
        \midrule
        \midrule
        \multirow{13}{*}{SAMSum} & \textsc{zero-shot} & - & - & 11.73 $\pm$ 0.14 & 1.357 $\pm$ 0.01 & 8.377 $\pm$ 0.19 & 9.331 $\pm$ 0.16\\
        & \textsc{baseline} & - & \text{\texttimes} & 32.95 $\pm$ 0.57 & 10.22 $\pm$ 0.15 & 25.83 $\pm$ 0.32 & 25.59 $\pm$ 0.29 \\
        & \textsc{cheating} & - & \text{\checkmark} & \textbf{36.36} $\pm$ 0.53 & \textbf{12.31} $\pm$ 0.14 & \textbf{28.41} $\pm$ 0.33 & \textbf{28.48} $\pm$ 0.33 \\
        \cline{2-8}
        \rule{0pt}{2.5ex}
        & \textsc{verbatim} & \text{\checkmark} & \text{\texttimes} & \textit{34.34} $\pm$ 0.45& \textit{10.76} $\pm$ 0.16& \textit{26.98} $\pm$ 0.40 & \textit{27.04} $\pm$ 0.38\\
        & \textsc{distribution} & \text{\checkmark} & \text{\texttimes} & 33.73 $\pm$ 0.51 & 10.32 $\pm$ 0.15 & 26.48 $\pm$ 0.31 & 26.56 $\pm$ 0.33\\
        & \textsc{masked} & \text{\checkmark} & \text{\texttimes} & 33.05 $\pm$ 0.46 & 10.46 $\pm$ 0.15 & 25.77 $\pm$ 0.30 & 25.81 $\pm$ 0.28\\
        & \textsc{noised} & \text{\checkmark} & \text{\texttimes} &  33.62 $\pm$ 0.43 & 10.27 $\pm$ 0.16 & 26.50 $\pm$ 0.37 & 26.49 $\pm$ 0.38\\
        & \textsc{reformatted} & \text{\checkmark} & \text{\texttimes} & 33.63 $\pm$ 0.39 & 10.25 $\pm$ 0.15 & 26.37 $\pm$ 0.31 & 26.46 $\pm$ 0.34\\
        
        \cline{2-8}
        \rule{0pt}{2.5ex}
        & \textsc{indomain-verbatim} & \text{\texttimes} & \text{\texttimes} & 33.61 $\pm$ 0.46 & 10.27 $\pm$ 0.14 & 26.39 $\pm$ 0.30 & 26.46 $\pm$ 0.35\\
        & \textsc{indomain-dist.} & \text{\texttimes} & \text{\texttimes} & 33.55 $\pm$ 0.42 & 10.26 $\pm$ 0.11 & 26.32 $\pm$ 0.33 & 26.44 $\pm$ 0.35 \\
        & \textsc{indomain-mask} & \text{\texttimes} & \text{\texttimes} & 32.87 $\pm$ 0.41 & 10.47 $\pm$ 0.12 & 25.74 $\pm$ 0.35 & 25.74 $\pm$ 0.31\\
        & \textsc{indomain-noise} & \text{\texttimes} & \text{\texttimes} & 33.67 $\pm$ 0.37 & 10.33 $\pm$ 0.13 & 26.38 $\pm$ 0.29 & 26.47 $\pm$ 0.28\\
        & \textsc{indomain-reform.} & \text{\texttimes} & \text{\texttimes} & 33.52 $\pm$ 0.34 & 10.24 $\pm$ 0.16 & 26.24 $\pm$ 0.28 & 26.34 $\pm$ 0.29\\
        \midrule
        \midrule
        \multirow{13}{*}{XSum} & \textsc{zero-shot} & - & - & 12.52 $\pm$ 0.11 & 2.059 $\pm$ 0.00 & 9.035 $\pm$ 0.16 & 10.27 $\pm$ 0.17\\
        & \textsc{baseline} & - & \text{\texttimes} & 26.28 $\pm$ 0.48 & 6.424 $\pm$ 0.12 & 19.80 $\pm$ 0.32 & 19.81 $\pm$ 0.33 \\
        & \textsc{cheating} & - & \text{\checkmark} & \textbf{29.87} $\pm$ 0.41 & \textbf{8.334} $\pm$ 0.13 & \textbf{22.97} $\pm$ 0.43& \textbf{22.98} $\pm$ 0.42 \\
        \cline{2-8}
        \rule{0pt}{2.5ex}
        & \textsc{verbatim} & \text{\checkmark} & \text{\texttimes} & 26.53 $\pm$ 0.51 & 6.820 $\pm$ 0.12 & 20.08 $\pm$ 0.33 & 20.03 $\pm$ 0.37\\
        & \textsc{distribution} & \text{\checkmark} & \text{\texttimes} & \textit{26.61} $\pm$ 0.42& \textit{6.885} $\pm$ 0.13 & \textit{20.12} $\pm$ 0.37& \textit{20.11} $\pm$ 0.37\\
        & \textsc{masked} & \text{\checkmark} & \text{\texttimes} & 24.50 $\pm$ 0.46 & 5.677 $\pm$ 0.12 & 18.16 $\pm$ 0.29 & 18.39 $\pm$ 0.31 \\
        & \textsc{noised} & \text{\checkmark} & \text{\texttimes} & 26.16 $\pm$ 0.39 & 6.599 $\pm$ 0.12 & 19.72 $\pm$ 0.35 & 19.72 $\pm$ 0.35 \\
        & \textsc{reformatted} & \text{\checkmark} & \text{\texttimes} & 26.27 $\pm$ 0.43 & 6.623 $\pm$ 0.12 & 19.86 $\pm$ 0.29 & 19.86 $\pm$ 0.30 \\
        
        \cline{2-8}
        \rule{0pt}{2.5ex}
        & \textsc{indomain-verbatim} & \text{\texttimes} & \text{\texttimes} & 26.43 $\pm$ 0.41 & 6.745 $\pm$ 0.14 & 19.99 $\pm$ 0.27 & 19.99 $\pm$ 0.40\\
        & \textsc{indomain-dist.} & \text{\texttimes} & \text{\texttimes} & 26.34 $\pm$ 0.40 & 6.666 $\pm$ 0.12 & 19.85 $\pm$ 0.32 & 19.85 $\pm$ 0.32\\
        & \textsc{indomain-mask} & \text{\texttimes} & \text{\texttimes} & 24.31 $\pm$ 0.39 & 5.521 $\pm$ 0.13 & 18.02 $\pm$ 0.29 & 18.04 $\pm$ 0.34\\
        & \textsc{indomain-noise} & \text{\texttimes} & \text{\texttimes} & 26.31 $\pm$ 0.46& 6.607 $\pm$ 0.11 & 19.80 $\pm$ 0.36& 19.81 $\pm$ 0.28\\
        & \textsc{indomain-reform.} & \text{\texttimes} & \text{\texttimes} & 25.29 $\pm$ 0.32 & 6.280 $\pm$ 0.12 & 19.04 $\pm$ 0.35 & 19.06 $\pm$ 0.30\\
        \bottomrule
    \end{tabular}
    }
     \caption{Results for all 13 models trained on XSum, SAMSum, and CNN/Daily Mail Datasets. The table showcases evaluation metrics, with the best-performing model scores bolded and the second best italicized.}
    \label{tab:full_summ}
\end{table*}

\definecolor{vermillion}{RGB}{214, 39, 40}
\definecolor{bluishgreen}{RGB}{44, 160, 44}
\definecolor{blue}{RGB}{100, 149, 237}

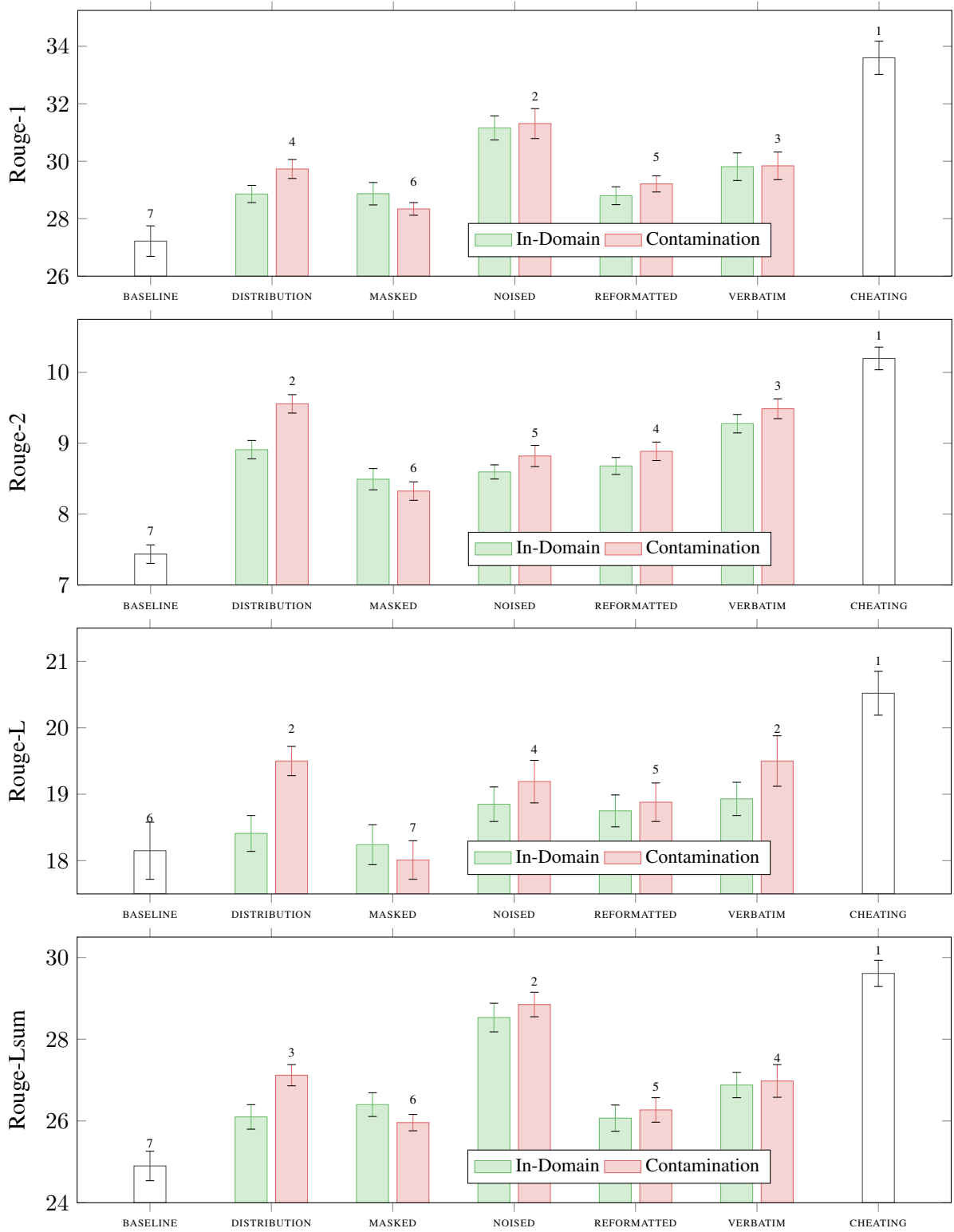
\begin{figure*}
\centering
\begin{tikzpicture}
\begin{axis}[
    ybar,
    width=\textwidth, 
    height=6cm, 
    bar width=15pt,
    ylabel={Rouge-1},
    xlabel style={font=\small, yshift=-0.75cm},
    area style,
    xtick=data,
    ymax=35.25,
    ymin=26,
    xticklabels={\textsc{baseline}, \textsc{distribution}, \textsc{masked}, \textsc{noised}, \textsc{reformatted}, \textsc{verbatim}, \textsc{cheating}},
    xticklabel style={align=center, font=\tiny, text width=3cm}, 
    legend style={at={(0.62,0.2)}, anchor=north,legend columns=-1, font=\footnotesize},
    nodes near coords, 
    nodes near coords align={vertical},
    nodes near coords style={font=\tiny, yshift=7pt},
    point meta=explicit symbolic,
    ]

    \addplot[fill=black!0, draw=black!70, point meta=explicit symbolic, bar width=0pt] coordinates {
        (2, 0)
        (3, 0)
        (4, 0)
        (5, 0)
        (6, 0)
        (7, 0)
        (8, 0)
    };

    \addplot[fill=bluishgreen!20, draw=bluishgreen!70, xshift=7.5pt, error bars/.cd,y dir=both,y explicit] coordinates {
        (3, 28.86) +- (0, 0.30) 
        (4, 28.87) +- (0, 0.39) 
        (5, 31.16) +- (0, 0.42) 
        (6, 28.80) +- (0, 0.31) 
        (7, 29.81) +- (0, 0.48) 
    };

     \addplot[fill=black!0, draw=black!70, error bars/.cd,y dir=both,y explicit] coordinates {
        (2, 27.22) +- (0, 0.53) [7]
        (8, 33.60) +- (0, 0.58) [1]
    };
    
    \addplot[fill=vermillion!20, draw=vermillion!70, xshift=-7.5pt, error bars/.cd,y dir=both,y explicit] coordinates {
        (3, 29.73) +- (0, 0.33) [4]
        (4, 28.34) +- (0, 0.22) [6]
        (5, 31.31) +- (0, 0.52) [2]
        (6, 29.21) +- (0, 0.28) [5]
        (7, 29.84) +- (0, 0.48) [3]
    };

    \addplot[fill=black!0, draw=black!70, point meta=explicit symbolic, bar width=0pt] coordinates {
        (2, 0)
        (3, 0)
        (4, 0)
        (5, 0)
        (6, 0)
        (7, 0)
        (8, 0)
    };

    \legend{,In-Domain, ,Contamination}
    
\end{axis}
\end{tikzpicture}

\begin{tikzpicture}
\begin{axis}[
    ybar,
    width=\textwidth, 
    height=6cm, 
    bar width=15pt,
    ylabel={Rouge-2},
    xlabel style={font=\small, yshift=-0.75cm},
    area style,
    xtick=data,
    ymax=10.75,
    ymin=7,
    xticklabels={\textsc{baseline}, \textsc{distribution}, \textsc{masked}, \textsc{noised}, \textsc{reformatted}, \textsc{verbatim}, \textsc{cheating}},
    xticklabel style={align=center, font=\tiny, text width=3cm}, 
    legend style={at={(0.62,0.2)}, anchor=north,legend columns=-1, font=\footnotesize},
    nodes near coords, 
    nodes near coords align={vertical},
    nodes near coords style={font=\tiny, yshift=5pt},
    point meta=explicit symbolic,
    ]

    \addplot[fill=black!0, draw=black!70, point meta=explicit symbolic, bar width=0pt] coordinates {
        (2, 0)
        (3, 0)
        (4, 0)
        (5, 0)
        (6, 0)
        (7, 0)
        (8, 0)
    };

    \addplot[fill=bluishgreen!20, draw=bluishgreen!70, xshift=7.5pt, error bars/.cd,y dir=both,y explicit] coordinates {
        (3, 8.91)  +- (0, 0.13) 
        (4, 8.493) +- (0, 0.15) 
        (5, 8.596) +- (0, 0.10) 
        (6, 8.68) +- (0, 0.12) 
        (7, 9.277) +- (0, 0.13) 
    };

     \addplot[fill=black!0, draw=black!70, error bars/.cd,y dir=both,y explicit] coordinates {
        (2, 7.436) +- (0, 0.13) [7]
        (8, 10.198) +- (0, 0.16) [1]
    };
    
    \addplot[fill=vermillion!20, draw=vermillion!70, xshift=-7.5, error bars/.cd,y dir=both,y explicit] coordinates {
        (3, 9.557) +- (0, 0.13) [2]
        (4, 8.326) +- (0, 0.13) [6]
        (5, 8.821) +- (0, 0.15) [5]
        (6, 8.887) +- (0, 0.13) [4]
        (7, 9.488) +- (0, 0.14) [3]
    };
    
    \addplot[fill=black!0, draw=black!70, point meta=explicit symbolic, bar width=0pt] coordinates {
        (2, 0)
        (3, 0)
        (4, 0)
        (5, 0)
        (6, 0)
        (7, 0)
        (8, 0)
    };

    \legend{,In-Domain, ,Contamination}
    
\end{axis}
\end{tikzpicture}

\begin{tikzpicture}
\begin{axis}[
    ybar,
    width=\textwidth, 
    height=6cm, 
    bar width=15pt,
    ylabel={Rouge-L},
    xlabel style={font=\small, yshift=-0.75cm},
    area style,
    xtick=data,
    ymax=21.5,
    ymin=17.5,
    xticklabels={\textsc{baseline}, \textsc{distribution}, \textsc{masked}, \textsc{noised}, \textsc{reformatted}, \textsc{verbatim}, \textsc{cheating}},
    xticklabel style={align=center, font=\tiny, text width=3cm}, 
    legend style={at={(0.62,0.2)}, anchor=north,legend columns=-1, font=\footnotesize},
    nodes near coords, 
    nodes near coords align={vertical},
    nodes near coords style={font=\tiny, yshift=9.5pt},
    point meta=explicit symbolic,
    ]

    \addplot[fill=black!0, draw=black!70, point meta=explicit symbolic, bar width=0pt] coordinates {
        (2, 0)
        (3, 0)
        (4, 0)
        (5, 0)
        (6, 0)
        (7, 0)
        (8, 0)
    };

    \addplot[fill=bluishgreen!20, draw=bluishgreen!70, xshift=7.5pt, error bars/.cd,y dir=both,y explicit] coordinates {
        (3, 18.41) +- (0, 0.27) 
        (4, 18.24) +- (0, 0.30) 
        (5, 18.85) +- (0, 0.26) 
        (6, 18.75) +- (0, 0.24) 
        (7, 18.93) +- (0, 0.25) 
    };

     \addplot[fill=black!0, draw=black!70, error bars/.cd,y dir=both,y explicit] coordinates {
        (2, 18.15) +- (0, 0.43) [6]
        (8, 20.52) +- (0, 0.33) [1]
    };
    
    \addplot[fill=vermillion!20, draw=vermillion!70, xshift=-7.5pt, error bars/.cd,y dir=both,y explicit] coordinates {
        (3, 19.50) +- (0, 0.22) [2]
        (4, 18.01) +- (0, 0.29) [7]
        (5, 19.19) +- (0, 0.32) [4]
        (6, 18.88) +- (0, 0.29) [5]
        (7, 19.50) +- (0, 0.38) [2]
    };
    
    \addplot[fill=black!0, draw=black!70, point meta=explicit symbolic, bar width=0pt] coordinates {
        (2, 0)
        (3, 0)
        (4, 0)
        (5, 0)
        (6, 0)
        (7, 0)
        (8, 0)
    };

    \legend{,In-Domain, ,Contamination}
    
\end{axis}
\end{tikzpicture}

\begin{tikzpicture}
\begin{axis}[
    ybar,
    width=\textwidth, 
    height=6cm, 
    bar width=15pt,
    ylabel={Rouge-Lsum},
    xlabel style={font=\small, yshift=-0.75cm},
    area style,
    xtick=data,
    ymax=30.5,
    ymin=24,
    xticklabels={\textsc{baseline}, \textsc{distribution}, \textsc{masked}, \textsc{noised}, \textsc{reformatted}, \textsc{verbatim}, \textsc{cheating}},
    xticklabel style={align=center, font=\tiny, text width=3cm}, 
    legend style={at={(0.62,0.2)}, anchor=north,legend columns=-1, font=\footnotesize},
    nodes near coords, 
    nodes near coords align={vertical},
    nodes near coords style={font=\tiny, yshift=5pt},
    point meta=explicit symbolic,
    ]

    \addplot[fill=black!0, draw=black!70, point meta=explicit symbolic, bar width=0pt] coordinates {
        (2, 0)
        (3, 0)
        (4, 0)
        (5, 0)
        (6, 0)
        (7, 0)
        (8, 0)
    };

    \addplot[fill=bluishgreen!20, draw=bluishgreen!70, xshift=7.5pt, error bars/.cd,y dir=both,y explicit] coordinates {
        (3, 26.1) +- (0, 0.30) 
        (4, 26.4) +- (0, 0.29) 
        (5, 28.53) +- (0, 0.35) 
        (6, 26.07) +- (0, 0.32) 
        (7, 26.88) +- (0, 0.31) 
    };

     \addplot[fill=black!0, draw=black!70, error bars/.cd,y dir=both,y explicit] coordinates {
        (2, 24.90) +- (0, 0.36) [7]
        (8, 29.61) +- (0, 0.32) [1]
    };

    \addplot[fill=vermillion!20, draw=vermillion!70, xshift=-7.5pt, error bars/.cd,y dir=both,y explicit] coordinates {
        (3, 27.12) +- (0, 0.26) [3]
        (4, 25.96) +- (0, 0.20) [6]
        (5, 28.85) +- (0, 0.30) [2]
        (6, 26.27) +- (0, 0.30) [5]
        (7, 26.98) +- (0, 0.40) [4]
    };

    \addplot[fill=black!0, draw=black!70, point meta=explicit symbolic, bar width=0pt] coordinates {
        (2, 0)
        (3, 0)
        (4, 0)
        (5, 0)
        (6, 0)
        (7, 0)
        (8, 0)
    };

    \legend{,In-Domain, ,Contamination}
    
\end{axis}
\end{tikzpicture}
\caption{Bar Chart of all CNN/Daily Mail models compared for each metric}
\label{fig:cdm_barchart}
\end{figure*}
\definecolor{vermillion}{RGB}{214, 39, 40}
\definecolor{bluishgreen}{RGB}{44, 160, 44}
\definecolor{blue}{RGB}{100, 149, 237}

\begin{figure*}
\centering
\begin{tikzpicture}
\begin{axis}[
    ybar,
    width=\textwidth, 
    height=6cm, 
    bar width=15pt,
    ylabel={Rouge-1},
    xlabel style={font=\small, yshift=-0.75cm},
    area style,
    xtick=data,
    ymax=37.5,
    ymin=32,
    xticklabels={ \textsc{baseline},  \textsc{distribution}, \textsc{masked}, \textsc{noised}, \textsc{reformatted}, \textsc{verbatim}, \textsc{cheating}},
    xticklabel style={align=center, font=\tiny, text width=3cm}, 
    legend style={at={(0.62,0.2)}, anchor=north,legend columns=-1, font=\footnotesize},
    nodes near coords, 
    nodes near coords align={vertical},
    nodes near coords style={font=\tiny, yshift=10pt},
    point meta=explicit symbolic,
    ]

    \addplot[fill=black!0, draw=black!70, point meta=explicit symbolic, bar width=0pt] coordinates {
        (2, 0)
        (3, 0)
        (4, 0)
        (5, 0)
        (6, 0)
        (7, 0)
        (8, 0)
    };
    
    \addplot[fill=bluishgreen!20, draw=bluishgreen!70,  xshift=7.5pt, error bars/.cd,y dir=both,y explicit] coordinates {
        (3, 33.55) +- (0, 0.42) 
        (4, 32.87) +- (0, 0.41) 
        (5, 33.67) +- (0, 0.37) 
        (6, 33.52) +- (0, 0.34) 
        (7, 33.61) +- (0, 0.46) 
    };

    \addplot[fill=black!0, draw=black!70, error bars/.cd,y dir=both,y explicit] coordinates {
        
        (2, 32.95) +- (0, 0.57) [7]
        (8, 36.36) +- (0, 0.53) [1]
    };

    \addplot[fill=vermillion!20, draw=vermillion!70, xshift=-7.5pt, error bars/.cd,y dir=both,y explicit] coordinates {
        (3, 33.73) +- (0, 0.51) [3]
        (4, 33.05) +- (0, 0.46) [6]
        (5, 33.62) +- (0, 0.43) [5]
        (6, 33.63) +- (0, 0.39) [4]
        (7, 34.34) +- (0, 0.45) [2]
    };

    \addplot[fill=black!0, draw=black!70, point meta=explicit symbolic, bar width=0pt] coordinates {
        (2, 0)
        (3, 0)
        (4, 0)
        (5, 0)
        (6, 0)
        (7, 0)
        (8, 0)
    };

    \legend{,In-Domain, ,Contamination}
    
\end{axis}
\end{tikzpicture}

\begin{tikzpicture}
\begin{axis}[
    ybar,
    width=\textwidth, 
    height=6cm, 
    bar width=15pt,
    ylabel={Rouge-2},
    xlabel style={font=\small, yshift=-0.75cm},
    area style,
    xtick=data,
    ymax=12.75,
    ymin=9.25,
    xticklabels={ \textsc{baseline}, \textsc{distribution}, \textsc{masked}, \textsc{noised}, \textsc{reformatted}, \textsc{verbatim}, \textsc{cheating}},
    xticklabel style={align=center, font=\tiny, text width=3cm}, 
    legend style={at={(0.62,0.2)}, anchor=north,legend columns=-1, font=\footnotesize},
    nodes near coords, 
    nodes near coords align={vertical},
    nodes near coords style={font=\tiny, yshift=4pt},
    point meta=explicit symbolic,
    ]

    \addplot[fill=black!0, draw=black!70, point meta=explicit symbolic, bar width=0pt] coordinates {
        (2, 0)
        (3, 0)
        (4, 0)
        (5, 0)
        (6, 0)
        (7, 0)
        (8, 0)
    };

    \addplot[fill=bluishgreen!20, draw=bluishgreen!70, point meta=explicit symbolic, xshift=7.5pt, error bars/.cd,y dir=both,y explicit] coordinates {
        (3, 10.26)  +- (0, 0.11) 
        (4, 10.47)  +- (0, 0.12) 
        (5, 10.33)  +- (0, 0.13) 
        (6, 10.24)  +- (0, 0.16) 
        (7, 10.27)  +- (0, 0.14) 
    };

    \addplot[fill=black!0, draw=black!70, error bars/.cd,y dir=both,y explicit] coordinates {
        (2, 10.22) +- (0, 0.15) [7]
        (8, 12.31) +- (0, 0.14) [1]
        
    };

    \addplot[fill=vermillion!20, draw=vermillion!70, xshift=-7.5pt, error bars/.cd,y dir=both,y explicit] coordinates {
        (3, 10.32) +- (0, 0.15) [4]
        (4, 10.46) +- (0, 0.15) [3]
        (5, 10.27) +- (0, 0.16) [5]
        (6, 10.25) +- (0, 0.15) [6]
        (7, 10.76) +- (0, 0.16) [2]
    };

    \addplot[fill=black!0, draw=black!70, point meta=explicit symbolic, bar width=0pt] coordinates {
        (2, 0)
        (3, 0)
        (4, 0)
        (5, 0)
        (6, 0)
        (7, 0)
        (8, 0)
    };

    \legend{,In-Domain, ,Contamination}
    
\end{axis}
\end{tikzpicture}

\begin{tikzpicture}
\begin{axis}[
    ybar,
    width=\textwidth, 
    height=6cm, 
    bar width=15pt,
    ylabel={Rouge-L},
    xlabel style={font=\small, yshift=-0.75cm},
    area style,
    xtick=data,
    ymax=29.5,
    ymin=25,
    xticklabels={\textsc{baseline},  \textsc{distribution}, \textsc{masked}, \textsc{noised}, \textsc{reformatted}, \textsc{verbatim}, \textsc{cheating}},
    xticklabel style={align=center, font=\tiny, text width=3cm}, 
    legend style={at={(0.62,0.2)}, anchor=north,legend columns=-1, font=\footnotesize},
    nodes near coords, 
    nodes near coords align={vertical},
    nodes near coords style={font=\tiny, yshift=8pt},
    point meta=explicit symbolic,
    ]

    \addplot[fill=black!0, draw=black!70, point meta=explicit symbolic, bar width=0pt] coordinates {
        (2, 0)
        (3, 0)
        (4, 0)
        (5, 0)
        (6, 0)
        (7, 0)
        (8, 0)
    };
    
    \addplot[fill=bluishgreen!20, draw=bluishgreen!70, xshift=7.5pt, error bars/.cd,y dir=both,y explicit] coordinates {
        (3, 26.32)  +- (0, 0.33)
        (4, 25.74)  +- (0, 0.35)
        (5, 26.38)  +- (0, 0.29)
        (6, 26.24)  +- (0, 0.28)
        (7, 26.50)  +- (0, 0.30)
    };

    \addplot[fill=black!0, draw=black!70, error bars/.cd,y dir=both,y explicit] coordinates {
        (2, 25.83) +- (0, 0.32) [6]
        (8, 28.41) +- (0, 0.33) [1]
        
    };

    \addplot[fill=vermillion!20, draw=vermillion!70, xshift=-7.5pt, error bars/.cd,y dir=both,y explicit] coordinates {
        (3, 26.48) +- (0, 0.31) [3]
        (4, 25.77) +- (0, 0.30) [7]
        (5, 26.39) +- (0, 0.37) [4]
        (6, 26.37) +- (0, 0.31) [5]
        (7, 26.98) +- (0, 0.40 )[2]
    };
    
   \addplot[fill=black!0, draw=black!70, point meta=explicit symbolic, bar width=0pt] coordinates {
        (2, 0)
        (3, 0)
        (4, 0)
        (5, 0)
        (6, 0)
        (7, 0)
        (8, 0)
    };

     \legend{,In-Domain, ,Contamination}
    
\end{axis}
\end{tikzpicture}

\begin{tikzpicture}
\begin{axis}[
    ybar,
    width=\textwidth, 
    height=6cm, 
    bar width=15pt,
    ylabel={Rouge-Lsum},
    xlabel style={font=\small, yshift=-0.75cm},
    area style,
    xtick=data,
    ymax=29.5,
    ymin=24.75,
    xticklabels={\textsc{baseline}, \textsc{distribution}, \textsc{masked}, \textsc{noised}, \textsc{reformatted}, \textsc{verbatim}, \textsc{cheating}},
    xticklabel style={align=center, font=\tiny, text width=3cm}, 
    legend style={at={(0.62,0.2)}, anchor=north,legend columns=-1, font=\footnotesize},
    nodes near coords, 
    nodes near coords align={vertical},
    nodes near coords style={font=\tiny, yshift=8pt},
    point meta=explicit symbolic,
    ]
    
    \addplot[fill=black!0, draw=black!70, point meta=explicit symbolic, bar width=0pt] coordinates {
        (2, 0)
        (3, 0)
        (4, 0)
        (5, 0)
        (6, 0)
        (7, 0)
        (8, 0)
    };
    
    \addplot[fill=bluishgreen!20, draw=bluishgreen!70, xshift=7.5pt, error bars/.cd,y dir=both,y explicit] coordinates {
        (3, 26.44)  +- (0, 0.35)
        (4, 25.74)  +- (0, 0.31)
        (5, 26.47)  +- (0, 0.28)
        (6, 26.34)  +- (0, 0.29)
        (7, 26.49)  +- (0, 0.35)
    };

    \addplot[fill=black!0, draw=black!70, error bars/.cd,y dir=both,y explicit] coordinates {
        (2, 25.59) +- (0, 0.29) [7]
        (8, 28.48) +- (0, 0.33)[1]
    };

    \addplot[fill=vermillion!20, draw=vermillion!70, xshift=-7.5pt, error bars/.cd,y dir=both,y explicit] coordinates {
        (3, 26.56) +- (0, 0.33) [3]
        (4, 25.81) +- (0, 0.28) [6]
        (5, 26.46) +- (0, 0.38) [5]
        (6, 26.46) +- (0, 0.34) [4]
        (7, 27.04) +- (0, 0.38) [2]
    };
    
    \addplot[fill=black!0, draw=black!70, point meta=explicit symbolic, bar width=0pt] coordinates {
        (2, 0)
        (3, 0)
        (4, 0)
        (5, 0)
        (6, 0)
        (7, 0)
        (8, 0)
    };

    \legend{,In-Domain, ,Contamination}
    
\end{axis}
\end{tikzpicture}
\caption{Bar Chart of all SAMSum models compared for each metric}
\label{fig:samsum_barchart}
\end{figure*}
\definecolor{vermillion}{RGB}{214, 39, 40}
\definecolor{bluishgreen}{RGB}{44, 160, 44}
\definecolor{blue}{RGB}{100, 149, 237}

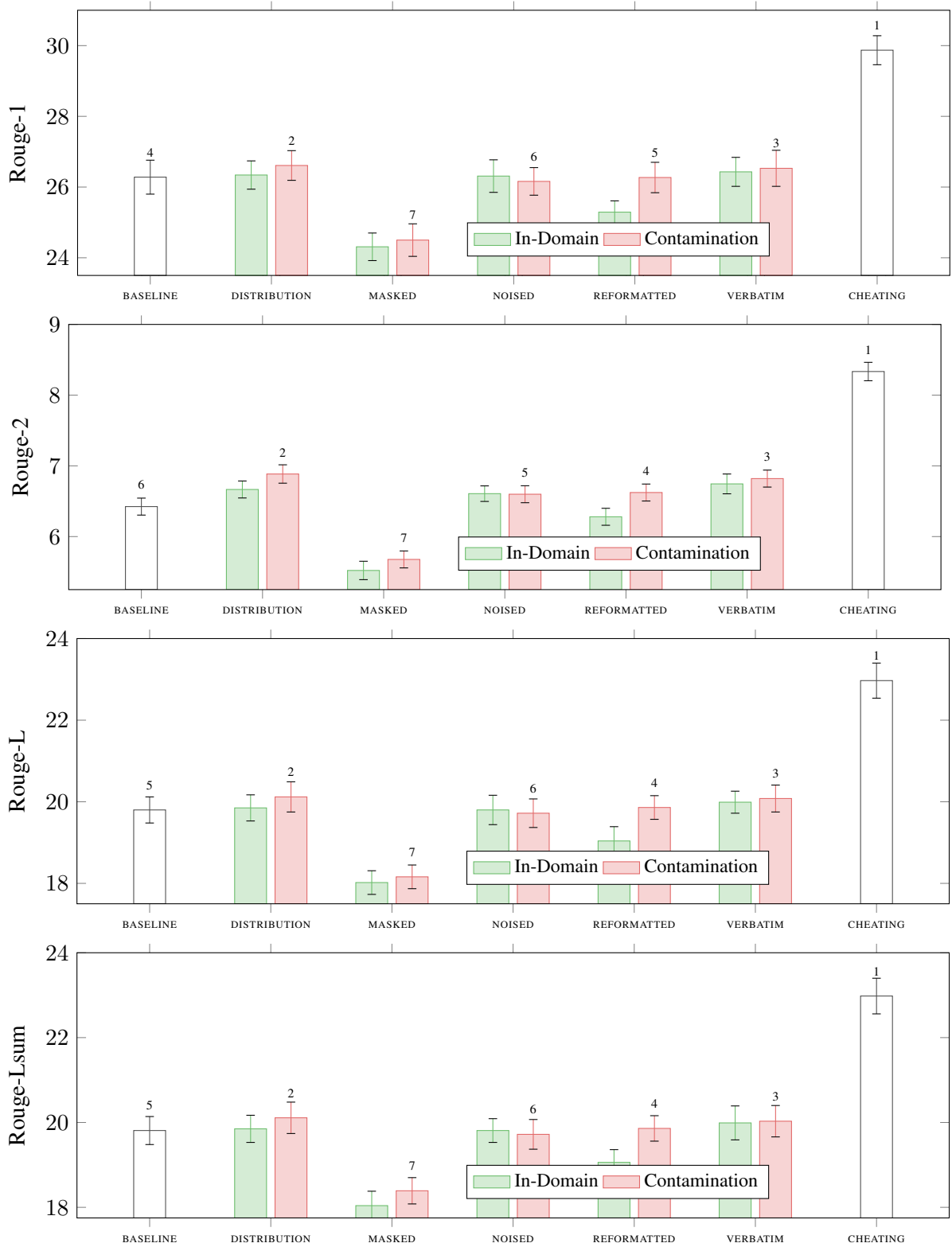
\begin{figure*}
\centering
\begin{tikzpicture}
\begin{axis}[
    ybar,
    width=\textwidth, 
    height=6cm, 
    bar width=15pt,
    ylabel={Rouge-1},
    xlabel style={font=\small, yshift=-0.75cm},
    area style,
    xtick=data,
    ymax=31,
    ymin=23.5,
    xticklabels={\textsc{baseline}, \textsc{distribution}, \textsc{masked}, \textsc{noised}, \textsc{reformatted}, \textsc{verbatim}, \textsc{cheating}},
    xticklabel style={align=center, font=\tiny, text width=3cm}, 
    legend style={at={(0.62,0.2)}, anchor=north,legend columns=-1, font=\footnotesize},
    nodes near coords, 
    nodes near coords align={vertical},
    nodes near coords style={font=\tiny, yshift=6pt},
    point meta=explicit symbolic,
    ]

    \addplot[fill=black!0, draw=black!70, point meta=explicit symbolic, bar width=0pt] coordinates {
        (2, 0)
        (3, 0)
        (4, 0)
        (5, 0)
        (6, 0)
        (7, 0)
        (8, 0)
    };

    \addplot[fill=bluishgreen!20, draw=bluishgreen!70, xshift=7.5pt, error bars/.cd,y dir=both,y explicit] coordinates {
        (3, 26.34) +- (0, 0.40)
        (4, 24.31) +- (0, 0.39)
        (5, 26.31) +- (0, 0.46)
        (6, 25.29) +- (0, 0.32)
        (7, 26.43) +- (0, 0.41)
    };

     \addplot[fill=black!0, draw=black!70, error bars/.cd,y dir=both,y explicit] coordinates {
        (2, 26.28) +- (0, 0.48) [4]
        (8, 29.87) +- (0, 0.41) [1]
    };
    
    \addplot[fill=vermillion!20, draw=vermillion!70, xshift=-7.5pt, error bars/.cd,y dir=both,y explicit] coordinates {
        (3, 26.61) +- (0, 0.42) [2]
        (4, 24.50) +- (0, 0.46) [7]
        (5, 26.16) +- (0, 0.39) [6]
        (6, 26.27) +- (0, 0.43) [5]
        (7, 26.53) +- (0, 0.51) [3]
    };

    \addplot[fill=black!0, draw=black!70, point meta=explicit symbolic, bar width=0pt] coordinates {
        (2, 0)
        (3, 0)
        (4, 0)
        (5, 0)
        (6, 0)
        (7, 0)
        (8, 0)
    };

    \legend{,In-Domain, ,Contamination}
    
\end{axis}
\end{tikzpicture}

\begin{tikzpicture}
\begin{axis}[
    ybar,
    width=\textwidth, 
    height=6cm, 
    bar width=15pt,
    ylabel={Rouge-2},
    xlabel style={font=\small, yshift=-0.75cm},
    area style,
    xtick=data,
    ymax=9,
    ymin=5.25,
    xticklabels={\textsc{baseline}, \textsc{distribution}, \textsc{masked}, \textsc{noised}, \textsc{reformatted}, \textsc{verbatim}, \textsc{cheating}},
    xticklabel style={align=center, font=\tiny, text width=3cm}, 
    legend style={at={(0.62,0.2)}, anchor=north,legend columns=-1, font=\footnotesize},
    nodes near coords, 
    nodes near coords align={vertical},
    nodes near coords style={font=\tiny, yshift=4.5pt},
    point meta=explicit symbolic,
    ]

    \addplot[fill=black!0, draw=black!70, point meta=explicit symbolic, bar width=0pt] coordinates {
        (2, 0)
        (3, 0)
        (4, 0)
        (5, 0)
        (6, 0)
        (7, 0)
        (8, 0)
    };

    \addplot[fill=bluishgreen!20, draw=bluishgreen!70, xshift=7.5pt, error bars/.cd,y dir=both,y explicit] coordinates {
        (3, 6.6661) +- (0, 0.12)
        (4, 5.521) +- (0, 0.13)
        (5, 6.607) +- (0, 0.11)
        (6, 6.28) +- (0, 0.12)
        (7, 6.745) +- (0, 0.14)
    };

     \addplot[fill=black!0, draw=black!70, error bars/.cd,y dir=both,y explicit] coordinates {
        (2, 6.424) +- (0, 0.12) [6]
        (8, 8.334) +- (0, 0.13) [1]
    };

    \addplot[fill=vermillion!20, draw=vermillion!70, xshift=-7.5pt, error bars/.cd,y dir=both,y explicit] coordinates {
        (3, 6.885) +- (0, 0.13) [2]
        (4, 5.677) +- (0, 0.12) [7]
        (5, 6.599) +- (0, 0.12) [5]
        (6, 6.623) +- (0, 0.12) [4]
        (7, 6.820) +- (0, 0.12) [3]
    };

    \addplot[fill=black!0, draw=black!70, point meta=explicit symbolic, bar width=0pt] coordinates {
        (2, 0)
        (3, 0)
        (4, 0)
        (5, 0)
        (6, 0)
        (7, 0)
        (8, 0)
    };

    \legend{,In-Domain, ,Contamination}
    
\end{axis}
\end{tikzpicture}

\begin{tikzpicture}
\begin{axis}[
    ybar,
    width=\textwidth, 
    height=6cm, 
    bar width=15pt,
    ylabel={Rouge-L},
    xlabel style={font=\small, yshift=-0.75cm},
    area style,
    xtick=data,
    ymax=24,
    ymin=17.5,
    xticklabels={\textsc{baseline}, \textsc{distribution}, \textsc{masked}, \textsc{noised}, \textsc{reformatted}, \textsc{verbatim}, \textsc{cheating}},
    xticklabel style={align=center, font=\tiny, text width=3cm}, 
    legend style={at={(0.62,0.2)}, anchor=north,legend columns=-1, font=\footnotesize},
    nodes near coords, 
    nodes near coords align={vertical},
    nodes near coords style={font=\tiny, yshift=6pt},
    point meta=explicit symbolic,
    ]

    \addplot[fill=black!0, draw=black!70, point meta=explicit symbolic, bar width=0pt] coordinates {
        (2, 0)
        (3, 0)
        (4, 0)
        (5, 0)
        (6, 0)
        (7, 0)
        (8, 0)
    };

    \addplot[fill=bluishgreen!20, draw=bluishgreen!70, xshift=7.5pt, error bars/.cd,y dir=both,y explicit] coordinates {
        (3, 19.85) +- (0, 0.32)
        (4, 18.02) +- (0, 0.29)
        (5, 19.80) +- (0, 0.36)
        (6, 19.04) +- (0, 0.35)
        (7, 19.99) +- (0, 0.27)
    };

     \addplot[fill=black!0, draw=black!70, error bars/.cd,y dir=both,y explicit] coordinates {
        (2, 19.80) +- (0, 0.32) [5]
        (8, 22.97) +- (0, 0.43) [1]
    };
    
    \addplot[fill=vermillion!20, draw=vermillion!70, xshift=-7.5pt, error bars/.cd,y dir=both,y explicit] coordinates {
        (3, 20.12) +- (0, 0.37) [2]
        (4, 18.16) +- (0, 0.29) [7]
        (5, 19.72) +- (0, 0.35) [6]
        (6, 19.86) +- (0, 0.29) [4]
        (7, 20.08) +- (0, 0.33) [3]
    };

    \addplot[fill=black!0, draw=black!70, point meta=explicit symbolic, bar width=0pt] coordinates {
        (2, 0)
        (3, 0)
        (4, 0)
        (5, 0)
        (6, 0)
        (7, 0)
        (8, 0)
    };
    \legend{,In-Domain, ,Contamination}
    
\end{axis}
\end{tikzpicture}

\begin{tikzpicture}
\begin{axis}[
    ybar,
    width=\textwidth, 
    height=6cm, 
    bar width=15pt,
    ylabel={Rouge-Lsum},
    xlabel style={font=\small, yshift=-0.75cm},
    area style,
    xtick=data,
    ymax=24,
    ymin=17.75,
    xticklabels={\textsc{baseline}, \textsc{distribution}, \textsc{masked}, \textsc{noised}, \textsc{reformatted}, \textsc{verbatim}, \textsc{cheating}},
    xticklabel style={align=center, font=\tiny, text width=3cm}, 
    legend style={at={(0.62,0.2)}, anchor=north,legend columns=-1, font=\footnotesize},
    nodes near coords, 
    nodes near coords align={vertical},
    nodes near coords style={font=\tiny, yshift=6pt},
    point meta=explicit symbolic,
    ]

    \addplot[fill=black!0, draw=black!70, point meta=explicit symbolic, bar width=0pt] coordinates {
        (2, 0)
        (3, 0)
        (4, 0)
        (5, 0)
        (6, 0)
        (7, 0)
        (8, 0)
    };

    \addplot[fill=bluishgreen!20, draw=bluishgreen!70, xshift=7.5pt, error bars/.cd,y dir=both,y explicit] coordinates {
        (3, 19.85) +- (0, 0.32)
        (4, 18.04) +- (0, 0.34)
        (5, 19.81) +- (0, 0.28)
        (6, 19.06) +- (0, 0.30)
        (7, 19.99) +- (0, 0.40)
    };

     \addplot[fill=black!0, draw=black!70, error bars/.cd,y dir=both,y explicit] coordinates {
        (2, 19.81) +- (0, 0.33) [5]
        (8, 22.98) +- (0, 0.42) [1]
    };
    
    \addplot[fill=vermillion!20, draw=vermillion!70, xshift=-7.5pt, error bars/.cd,y dir=both,y explicit] coordinates {
        (3, 20.11) +- (0, 0.37) [2]
        (4, 18.39) +- (0, 0.31) [7]
        (5, 19.72) +- (0, 0.35) [6]
        (6, 19.86) +- (0, 0.30) [4]
        (7, 20.03) +- (0, 0.37) [3]
    };

    \addplot[fill=black!0, draw=black!70, point meta=explicit symbolic, bar width=0pt] coordinates {
        (2, 0)
        (3, 0)
        (4, 0)
        (5, 0)
        (6, 0)
        (7, 0)
        (8, 0)
    };

    \legend{,In-Domain, ,Contamination}
    
\end{axis}
\end{tikzpicture}
\caption{Bar Chart of all XSum models compared for each metric}
\label{fig:xsum_barchart}
\end{figure*}
\clearpage

\section{Full results for Question Answering Case Study}
We present the full results of the QA case study. For each setting and dataset, we have included a table of the exact match and f1 metrics along with their standard deviations. The data is also presented through a series of bar charts for easier interpretability of the results for the reader. Standard deviations are measured over the results of the 3 models trained on random shuffles of the data.
\label{app:full_results_qa}
\begin{table*}[ht]
    \centering
    \footnotesize
    \begin{tabular}{l|c|>{\centering\arraybackslash}p{2.5cm}|>{\centering\arraybackslash}p{2.5cm}|c|c|cr}
        \toprule
        \multirow{2}{*}{Dataset} & \multirow{2}{*}{Model} & Contaminated & Contaminated & \multirow{2}{*}{Exact Match} & \multirow{2}{*}{F1 Score} \\
        & & Pretraining Data & Fine-tuning Data & & \\
        \midrule
        \multirow{15}{*}{SQuAD} & \textsc{zero-shot} & - & - & 1.178 $\pm$ 0.11 & 4.180 $\pm$ 0.22\\
        & \textsc{baseline} & - & \text{\texttimes} & 41.76 $\pm$ 1.01 & 55.72 $\pm$ 0.85\\
        & \textsc{cheating} & - & \text{\checkmark} & \textbf{55.73} $\pm$ 0.94 & \textbf{66.47} $\pm$ 0.80\\
        \cline{2-6}
        \rule{0pt}{2.5ex}
        & \textsc{verbatim} & \text{\checkmark} & \text{\texttimes} & 53.38 $\pm$ 0.94 & 65.07 $\pm$ 0.96\\
        & \textsc{\textsc{distribution}} & \text{\checkmark} & \text{\texttimes} & 52.76 $\pm$ 0.89 & 64.92 $\pm$ 0.88\\
        & \textsc{masked} & \text{\checkmark} & \text{\texttimes} & 38.77 $\pm$ 0.96 & 51.93 $\pm$ 0.78 \\
        & \textsc{noised} & \text{\checkmark} & \text{\texttimes} & 52.72 $\pm$ 0.89 & 64.65 $\pm$ 0.89 \\
        & \textsc{reformatted} & \text{\checkmark} & \text{\texttimes} & 48.08 $\pm$ 0.91 & 61.85 $\pm$ 0.94 \\
        & \textsc{augmented} & \text{\checkmark} & \text{\texttimes} & \textit{53.58} $\pm$ 0.98 & \textit{65.51} $\pm$ 0.90 \\
        \cline{2-6}
        \rule{0pt}{2.5ex}
        & \textsc{indomain-verbatim} & \text{\texttimes} & \text{\texttimes} & 52.44 $\pm$ 0.89 & 64.52 $\pm$ 0.92 \\
        & \textsc{indomain-dist.} & \text{\texttimes} & \text{\texttimes} & 51.90 $\pm$ 0.91 & 64.43 $\pm$ 0.87\\
        & \textsc{indomain-mask} & \text{\texttimes} & \text{\texttimes} & 44.62 $\pm$ 0.93 & 58.95 $\pm$ 1.00\\
        & \textsc{indomain-noise} & \text{\texttimes} & \text{\texttimes} & 50.63 $\pm$ 0.85 & 63.60 $\pm$ 0.86\\
        & \textsc{indomain-reform.} & \text{\texttimes} & \text{\texttimes} & 51.30 $\pm$ 0.95 & 63.72 $\pm$ 0.95\\
        & \textsc{indomain-augment} & \text{\texttimes} & \text{\texttimes} & 52.94 $\pm$ 0.94 & 64.24 $\pm$ 0.89\\
        \midrule
        \midrule
        \multirow{15}{*}{CBT} & \textsc{zero-shot} & - & - & 1.192 $\pm$ 0.12 & 3.290 $\pm$ 0.21 \\
        & \textsc{baseline} & - & \text{\texttimes} & 19.41 $\pm$ 0.99 & 19.84 $\pm$ 0.90 \\
        & \textsc{cheating} & - & \text{\checkmark} & \textbf{54.27} $\pm$ 0.85 & \textbf{56.39} $\pm$ 0.96\\
        \cline{2-6}
        \rule{0pt}{2.5ex}
        & \textsc{verbatim} & \text{\checkmark} & \text{\texttimes} & \textit{52.06} $\pm$ 0.88 & \textit{53.91} $\pm$ 0.89 \\
        & \textsc{\textsc{distribution}} & \text{\checkmark} & \text{\texttimes} & 50.82 $\pm$ 0.97 & 51.21 $\pm$ 0.97 \\
        & \textsc{masked} & \text{\checkmark} & \text{\texttimes} & 46.51 $\pm$ 0.84 & 47.43 $\pm$ 0.93 \\
        & \textsc{noised} & \text{\checkmark} & \text{\texttimes} & 49.59 $\pm$ 0.86 & 50.44 $\pm$ 0.96 \\
        & \textsc{reformatted} & \text{\checkmark} & \text{\texttimes} & 51.46 $\pm$ 0.93 & 52.96 $\pm$ 0.86 \\
        & \textsc{augmented} & \text{\checkmark} & \text{\texttimes} & 49.09 $\pm$ 1.00 & 50.32 $\pm$ 0.89\\
        \cline{2-6}
        \rule{0pt}{2.5ex}
        & \textsc{indomain-verbatim} & \text{\texttimes} & \text{\texttimes} & 44.19 $\pm$ 0.87 & 45.06 $\pm$ 0.96\\
        & \textsc{indomain-dist.} & \text{\texttimes} & \text{\texttimes} & 42.85 $\pm$ 0.92 & 46.06 $\pm$ 0.90 \\
        & \textsc{indomain-mask} & \text{\texttimes} & \text{\texttimes} & 40.77 $\pm$ 0.96 & 40.18 $\pm$ 0.93\\
        & \textsc{indomain-noise} & \text{\texttimes} & \text{\texttimes} & 49.02 $\pm$ 0.97 & 49.11 $\pm$ 0.98\\
        & \textsc{indomain-reform.} & \text{\texttimes} & \text{\texttimes} & 50.01 $\pm$ 0.86 & 51.12 $\pm$ 0.86\\
        & \textsc{indomain-augment} & \text{\texttimes} & \text{\texttimes} & 50.46 $\pm$ 0.93 & 51.62 $\pm$ 0.84\\
        \bottomrule
    \end{tabular}
     \caption{Results for all 15 models trained on the SQuAD and CBT dataset. The table showcases evaluation metrics, with the best-performing model scores bolded and the
second best italicized.}
    \label{tab:full_qa}
\end{table*}
\definecolor{vermillion}{RGB}{214, 39, 40}
\definecolor{bluishgreen}{RGB}{44, 160, 44}
\definecolor{blue}{RGB}{100, 149, 237}

\begin{figure*}
\centering
\begin{tikzpicture}
\begin{axis}[
    ybar,
    width=\textwidth, 
    height=6cm, 
    bar width=15pt,
    ymin=37,
    ymax=58.5,
    ylabel={Exact Match},
    xlabel style={font=\small, yshift=-0.75cm},
    area style,
    xtick=data,
    xticklabels={\textsc{baseline},  \textsc{distribution}, \textsc{masked}, \textsc{noised}, \textsc{reformatted}, \textsc{augmented}, \textsc{verbatim}, \textsc{cheating}},
    xticklabel style={align=center, font=\tiny, text width=3cm}, 
    legend style={at={(0.62,0.2)}, anchor=north,legend columns=-1, font=\footnotesize},
    nodes near coords, 
    nodes near coords align={vertical},
    nodes near coords style={font=\tiny, yshift=4pt},
    point meta=explicit symbolic,
    ]

    \addplot[fill=black!0, draw=black!70, point meta=explicit symbolic, bar width=0pt] coordinates {
        (2, 0)
        (3, 0)
        (4, 0)
        (5, 0)
        (6, 0)
        (7, 0)
        (8, 0)
        (9, 0)
    };
    
    \addplot[fill=bluishgreen!20, draw=bluishgreen!70, xshift=7.5pt, error bars/.cd,y dir=both,y explicit] coordinates {
        (3, 51.895) +- (0, 0.91)
        (4, 44.615) +- (0, 0.93)
        (5, 50.625) +- (0, 0.85)
        (6, 51.3) +- (0, 0.95)
        (7, 52.94) +- (0, 0.94)
        (8, 52.442) +- (0, 0.89)
    };

     \addplot[fill=black!0, draw=black!70, error bars/.cd,y dir=both,y explicit] coordinates {
        (2, 41.768) +- (0, 1.01) [7]
        (9, 55.731) +- (0, 0.94) [1]
    };

    \addplot[fill=vermillion!20, draw=vermillion!70, xshift=-7.5pt, error bars/.cd,y dir=both,y explicit] coordinates {
        (3, 52.768) +- (0, 0.89) [4]
        (4, 38.779) +- (0, 0.96) [8]
        (5, 52.726) +- (0, 0.89) [5]
        (6, 48.081) +- (0, 0.91) [6]
        (7, 53.589) +- (0, 0.98) [2]
        (8, 53.389) +- (0, 0.94) [3]
    };

    \addplot[fill=black!0, draw=black!70, point meta=explicit symbolic, bar width=0pt] coordinates {
        (2, 0)
        (3, 0)
        (4, 0)
        (5, 0)
        (6, 0)
        (7, 0)
        (8, 0)
        (9,0)
    };
    
    \legend{,In-Domain, ,Contamination}
    
\end{axis}
\end{tikzpicture}

\begin{tikzpicture}
\begin{axis}[
    ybar,
    width=\textwidth, 
    height=6cm, 
    bar width=15pt,
    ymin=50,
    ymax=69.5,
    ylabel={F1 Score},
    xlabel style={font=\small, yshift=-0.75cm},
    area style,
    xtick=data,
    xticklabels={\textsc{baseline},  \textsc{distribution}, \textsc{masked}, \textsc{noised}, \textsc{reformatted}, \textsc{augmented}, \textsc{verbatim}, \textsc{cheating}},
    xticklabel style={align=center, font=\tiny, text width=3cm}, 
    legend style={at={(0.62,0.2)}, anchor=north,legend columns=-1, font=\footnotesize},
    nodes near coords, 
    nodes near coords align={vertical},
    nodes near coords style={font=\tiny, yshift=5pt},
    point meta=explicit symbolic,
    ]

    \addplot[fill=black!0, draw=black!70, point meta=explicit symbolic, bar width=0pt] coordinates {
        (2, 0)
        (3, 0)
        (4, 0)
        (5, 0)
        (6, 0)
        (7, 0)
        (8, 0)
        (9,0)
    };
    
    \addplot[fill=bluishgreen!20, draw=bluishgreen!70, xshift=7.5pt, error bars/.cd,y dir=both,y explicit] coordinates {
        (3, 64.425) +- (0, 0.87)
        (4, 58.947) +- (0, 1.00)
        (5, 63.599) +- (0, 0.86)
        (6, 63.72) +- (0, 0.95)
        (7, 64.241) +- (0, 0.89)
        (8, 64.528) +- (0, 0.92)
    };

    \addplot[fill=black!0, draw=black!70, error bars/.cd,y dir=both,y explicit] coordinates {
        (2, 55.729) +- (0, 0.85) [7]
        (9, 66.472) +- (0, 0.80) [1]
    };

    \addplot[fill=vermillion!20, draw=vermillion!70, xshift=-7.5pt, error bars/.cd,y dir=both,y explicit] coordinates {
        (3, 64.927) +- (0, 0.96) [4]
        (4, 51.936) +- (0, 0.88) [8]
        (5, 64.657) +- (0, 0.78) [5]
        (6, 61.851) +- (0, 0.89) [6]
        (7, 65.517) +- (0, 0.94) [2]
        (8, 65.076) +- (0, 0.90) [3]
    };

    \addplot[fill=black!0, draw=black!70, point meta=explicit symbolic, bar width=0pt] coordinates {
        (2, 0)
        (3, 0)
        (4, 0)
        (5, 0)
        (6, 0)
        (7, 0)
        (8, 0)
        (9,0)
    };
    
    \legend{,In-Domain, ,Contamination}
    
\end{axis}
\end{tikzpicture}
\caption{Bar Chart of all SQuAD models compared for each metric}
\label{fig:squad_barchart}
\end{figure*}

\definecolor{vermillion}{RGB}{214, 39, 40}
\definecolor{bluishgreen}{RGB}{44, 160, 44}
\definecolor{blue}{RGB}{100, 149, 237}

\begin{figure*}
\centering
\begin{tikzpicture}
\begin{axis}[
    ybar,
    width=\textwidth, 
    height=6cm, 
    bar width=15pt,
    ymin=17,
    ymax=59,
    ylabel={Exact Match},
    xlabel style={font=\small, yshift=-0.75cm},
    area style,
    xtick=data,
    xticklabels={\textsc{baseline}, \textsc{distribution}, \textsc{masked}, \textsc{noised}, \textsc{reformatted}, \textsc{augmented}, \textsc{verbatim}, \textsc{cheating}},
    xticklabel style={align=center, font=\tiny, text width=3cm}, 
    legend style={at={(0.62,0.2)}, anchor=north,legend columns=-1, font=\footnotesize},
    nodes near coords, 
    nodes near coords align={vertical},
    nodes near coords style={font=\tiny, yshift=3pt},
    point meta=explicit symbolic,
    ]

    \addplot[fill=black!0, draw=black!70, point meta=explicit symbolic, bar width=0pt] coordinates {
        (2, 0)
        (3, 0)
        (4, 0)
        (5, 0)
        (6, 0)
        (7, 0)
        (8, 0)
        (9, 0)
    };

    \addplot[fill=bluishgreen!20, draw=bluishgreen!70, xshift=7.5pt, error bars/.cd,y dir=both,y explicit] coordinates {
        (3, 42.85) +- (0, 0.92)
        (4, 40.77) +- (0, 0.96)
        (5, 49.02) +- (0, 0.97)
        (6, 50.01) +- (0, 0.86)
        (7, 50.46) +- (0, 0.93)
        (8, 44.19) +- (0, 0.87)
    };

     \addplot[fill=black!0, draw=black!70, error bars/.cd,y dir=both,y explicit] coordinates {
       (2, 19.41) +- (0, 0.99) [8]
       (9, 54.27) +- (0, 0.85) [1]
    };

     \addplot[fill=vermillion!20, draw=vermillion!70, xshift=-7.5pt, error bars/.cd,y dir=both,y explicit] coordinates {
        (3, 50.82) +- (0, 0.97) [4]
        (4, 46.51) +- (0, 0.84) [7]
        (5, 49.59) +- (0, 0.86) [5]
        (6, 51.46) +- (0, 0.93) [3]
        (7, 49.09) +- (0, 1.00) [6]
        (8, 52.06) +- (0, 0.88) [2]
    };

    \addplot[fill=black!0, draw=black!70] coordinates {
        (2, 0)
        (3, 0)
        (4, 0)
        (5, 0)
        (6, 0)
        (7, 0)
        (8, 0)
        (9, 0)
    };
    
    \legend{,In-Domain, ,Contamination}
    
\end{axis}
\end{tikzpicture}

\begin{tikzpicture}
\begin{axis}[
    ybar,
    width=\textwidth, 
    height=6cm, 
    bar width=15pt,
    ymin=17,
    ymax = 60,
    ylabel={F1 Score},
    xlabel style={font=\small, yshift=-0.75cm},
    area style,
    xtick=data,
    xticklabels={\textsc{baseline}, \textsc{distribution}, \textsc{masked}, \textsc{noised}, \textsc{reformatted}, \textsc{augmented}, \textsc{verbatim}, \textsc{cheating}},
    xticklabel style={align=center, font=\tiny, text width=3cm}, 
    legend style={at={(0.62,0.2)}, anchor=north,legend columns=-1, font=\footnotesize},
    nodes near coords, 
    nodes near coords align={vertical},
    nodes near coords style={font=\tiny, yshift=1pt},
    point meta=explicit symbolic,
    ]

    \addplot[fill=black!0, draw=black!70,] coordinates {
        (2, 0)
        (3, 0)
        (4, 0)
        (5, 0)
        (6, 0)
        (7, 0)
        (8, 0)
        (9,0)
    };
    
    \addplot[fill=bluishgreen!20, draw=bluishgreen!70, xshift=7.5pt, error bars/.cd,y dir=both,y explicit] coordinates {
        (3, 46.06) +- (0, 0.90)
        (4, 40.18) +- (0, 0.93)
        (5, 49.11) +- (0, 0.98)
        (6, 51.12) +- (0, 0.86)
        (7, 51.62) +- (0, 0.84)
        (8, 45.06) +- (0, 0.96)
    };

    \addplot[fill=black!0, draw=black!70, error bars/.cd,y dir=both,y explicit] coordinates {
        (2, 19.84) +- (0, 0.90) [8]
        (9, 56.39) +- (0, 0.96) [1]
    };

    \addplot[fill=vermillion!20, draw=vermillion!70, xshift=-7.5pt, error bars/.cd,y dir=both,y explicit] coordinates {
        (3, 51.21) +- (0, 0.97) [4]
        (4, 47.43) +- (0, 0.93) [7]
        (5, 50.44) +- (0, 0.96) [5]
        (6, 52.96) +- (0, 0.86) [3]
        (7, 50.32) +- (0, 0.89) [6]
        (8, 53.91) +- (0, 0.89) [2]
    };

    \addplot[fill=black!0, draw=black!70,] coordinates {
        (2, 0)
        (3, 0)
        (4, 0)
        (5, 0)
        (6, 0)
        (7, 0)
        (8, 0)
        (9,0)
    };
    
    \legend{,In-Domain, ,Contamination}
    
\end{axis}
\end{tikzpicture}
\caption{Bar Chart of all CBT models compared for each metric}
\label{fig:cbt_barchart}
\end{figure*}

\clearpage

\section{Examples for each contamination type}
 We provide examples of each of the functions from the different contamination types we are testing, applied to a sample from each dataset from the case studies.
\label{app:examples}

\begin{table*}[!htb]
  \centering
  \footnotesize
  \begin{tabular}{c|l}
    \toprule
    \multirow{7}{*}{Sample} & Conversation: \\
    & Anita: I'm at the station in Bologna \\
    & Jenny: No problems so far? \\
    & Anita: no, everything's going smoothly\\
    & Tomy: good!\\
    & \\
    & Summary: Anita is at Bologna station.\\
    \midrule
    \multirow{11}{*}{Distribution} & \textit{$\langle$ some open web text $\rangle$} \\
    & \\
    & Conversation: \\
    & Anita: I'm at the station in Bologna \\
    & Jenny: No problems so far? \\
    & Anita: no, everything's going smoothly\\
    & Tomy: good!\\
    & \\
    & Summary: Anita is at Bologna station.\\
    & \\
    & \textit{$\langle$ some more open web text $\rangle$} \\
    \midrule
    Masking & Summary: Anita is at Bologna station.\\
    \midrule
    \multirow{8}{*}{Noising} & Conversation: \\
    & Anita: I'm at the station in Bologna \\
    & Jenny: No problems so far? \\
    & Anita: no, everything's going smoothly\\
    & Tomy: good!\\
    & \\
    & Summary: Anita confirms her location at the Bologna station to Jenny and Tomy, \\
    & reassuring them that everything is running smoothly.\\
    \midrule
    \multirow{7}{*}{Reformatting} & Summary: Anita is at Bologna station.\\
    & \\
    & Conversation: \\
    & Anita: I'm at the station in Bologna \\
    & Jenny: No problems so far? \\
    & Anita: no, everything's going smoothly\\
    & Tomy: good!\\
    \bottomrule
  \end{tabular}
  \caption{Applying the different contamination techniques to a sample from the SAMSum dataset.}
  \label{tab:summ_example}
\end{table*}

\begin{table*} [!htb]
  \centering
  \scriptsize
  \begin{tabular}{c|l}
    \toprule
    \multirow{8}{*}{Sample} & Context: \\
    & The Bey Hive is the name given to Beyoncé's fan base. Fans were previously titled ``The Beyontourage'',\\
    & (a portmanteau of Beyoncé and entourage). The name Bey Hive derives from the word beehive, purposely misspelled \\
    & to resemble her first name, and was penned by fans after petitions on the online social networking service Twitter and online \\
    & news reports during competitions. \\
    & \\
    & Question: Beyonce has a fan base that is referred to as what? \\
    & Answer: The Bey Hive\\
    \midrule
    \multirow{12}{*}{Distribution} & \textit{$\langle$ some open web text $\rangle$} \\
    & \\
    & Context: \\
    & The Bey Hive is the name given to Beyoncé's fan base. Fans were previously titled ``The Beyontourage'',\\
    & (a portmanteau of Beyoncé and entourage). The name Bey Hive derives from the word beehive, purposely misspelled \\
    & to resemble her first name, and was penned by fans after petitions on the online social networking service Twitter and online \\
    & news reports during competitions. \\
    & \\
    & Question: Beyonce has a fan base that is referred to as what? \\
    & Answer: The Bey Hive\\
    & \\
    & \textit{$\langle$ some more open web text $\rangle$} \\
    \midrule
    \multirow{2}{*}{Masking} & Question: Beyonce has a fan base that is referred to as what? \\
    & Answer: The Bey Hive\\
    \midrule
    \multirow{8}{*}{Noising} & Context: \\
    & The Bey Hive is the name given to Beyoncé's fan base. Fans were previously titled ``The Beyontourage'',\\
    & (a portmanteau of Beyoncé and entourage). The name Bey Hive derives from the word beehive, purposely misspelled \\
    & to resemble her first name, and was penned by fans after petitions on the online social networking service Twitter and online \\
    & news reports during competitions. \\
    & \\
    & Question: Beyonce has a fan base that is referred to as what? \\
    & Answer: Bey Hive\\
    \midrule
    \multirow{14}{*}{Reformatting} & Context: \\
    & The Bey Hive is the name given to Beyoncé's fan base. Fans were previously titled ``The Beyontourage'',\\
    & (a portmanteau of Beyoncé and entourage). The name Bey Hive derives from the word beehive, purposely misspelled \\
    & to resemble her first name, and was penned by fans after petitions on the online social networking service Twitter and online \\
    & news reports during competitions. \\
    & \\
    & Question: Beyonce has a fan base that is referred to as what? \\
    & Options: \\
    & A) The Beehivers\\
    & B) The Bey Hive\\
    & C) The Beyontourage\\
    & D) The Bey Flock\\
    & \\
    & Answer: The Bey Hive\\
    \midrule
    \multirow{8}{*}{Augmenting} & Context: \\
    & The Bey Hive is the name given to Beyoncé's fan base. Fans were previously titled ``The Beyontourage'',\\
    & (a portmanteau of Beyoncé and entourage). The name Bey Hive derives from the word beehive, purposely misspelled \\
    & to resemble her first name, and was penned by fans after petitions on the online social networking service Twitter and online \\
    & news reports during competitions. This fervent fan base actively engages with Beyoncé's music, performances, and \\
    & philanthropic endeavors.\\
    & \\
    & Question: Beyonce has a fan base that is referred to as what? \\
    & Answer: The Bey Hive\\
    \bottomrule
  \end{tabular}
  \caption{Applying the different contamination techniques to a sample from the SQuAD dataset.}
  \label{tab:qa_example}
\end{table*}

\end{document}